\documentclass[10pt,twocolumn,letterpaper]{article}

\usepackage{iccv23/iccv}
\usepackage{times}
\usepackage{epsfig}
\usepackage{graphicx}
\usepackage{amsmath}
\usepackage{amssymb}

\usepackage[pagebackref=true,breaklinks=true,letterpaper=true,colorlinks,bookmarks=false]{hyperref}

\usepackage{cleveref} 

\iccvfinalcopy 

\def\iccvPaperID{11065} 

\ificcvfinal\fi

\begin{document}

\title{On Evaluating the Adversarial Robustness of Semantic Segmentation Models} 

\author{Levente Halmosi \qquad  Márk Jelasity\\
University of Szeged\\
Szeged, Hungary\\
{\tt\small jelasity@inf.u-szeged.hu}
}

\maketitle
\ificcvfinal\thispagestyle{empty}\fi

\begin{abstract}
Achieving robustness against adversarial input perturbation
is an important and intriguing problem in machine learning.
In the area of semantic image segmentation, a number of adversarial training
approaches have been proposed as a defense against adversarial perturbation,
but the methodology of evaluating the robustness of the
models is still lacking, compared to image classification.
Here, we demonstrate that, just like in image classification, it is
important to evaluate the models over several different and hard attacks.
We propose a set of gradient based iterative attacks and show that
it is essential to perform a large number of iterations.
We include attacks against the internal representations of the models as well.
We apply two types of attacks: maximizing the error with a bounded perturbation,
and minimizing the perturbation for a given level of error.
Using this set of attacks, we show for the first time
that a number of models in previous work that are claimed to be robust are in fact
not robust at all.
We then evaluate simple adversarial training algorithms that produce reasonably robust models
even under our set of strong attacks.
Our results indicate that a key design decision to achieve any robustness
is to use only adversarial examples during training.
However, this introduces a trade-off between robustness and accuracy.
\end{abstract}

\section{Introduction}
\label{sec:intro}


\begin{figure}[tb]
\centering
\includegraphics[width=0.4\textwidth]{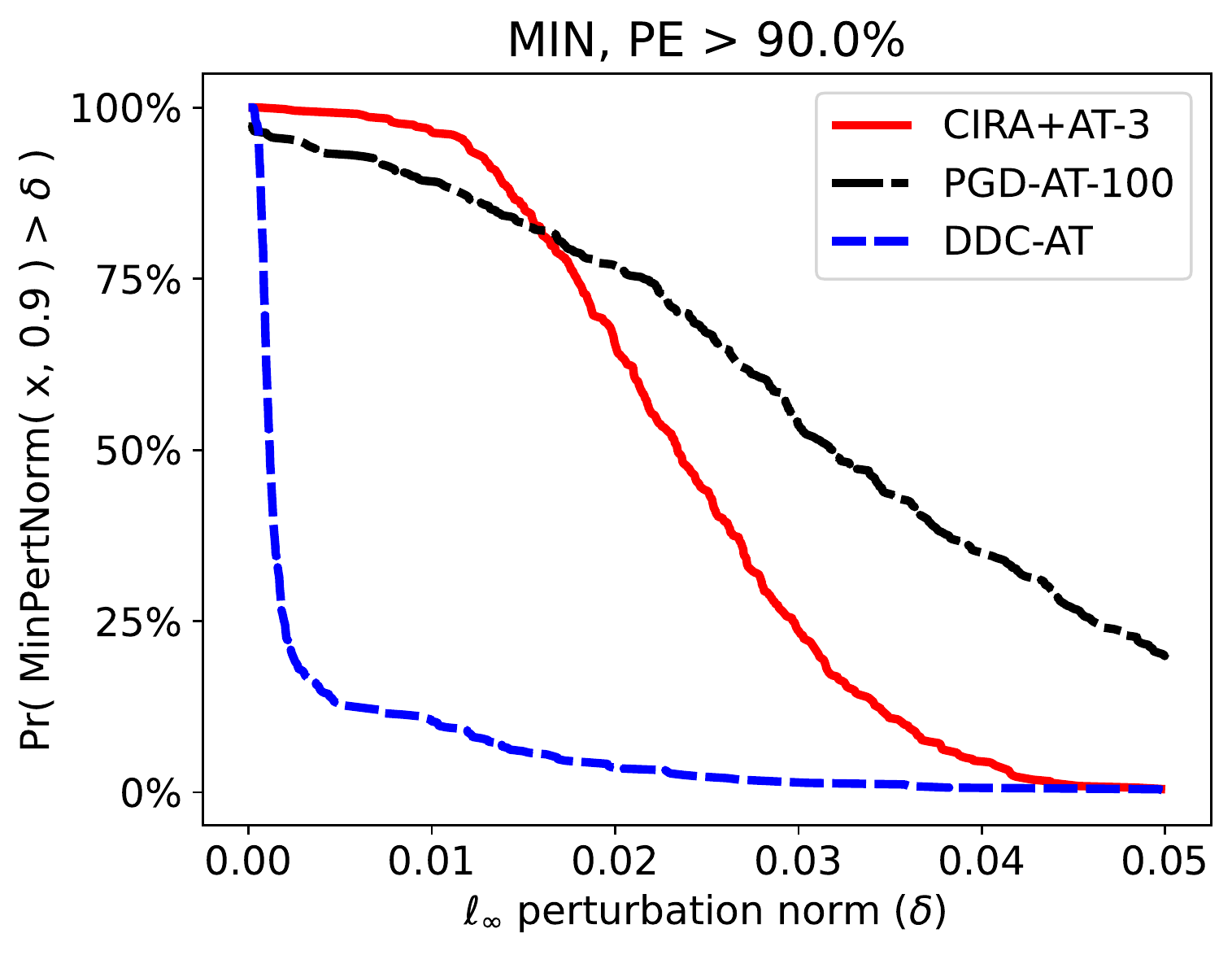}
\caption{Empirical probability that the minimal perturbation size
(in $\ell_\infty$ norm)
to achieve 90\% pixel error on a random input $x$ is larger than a given value.
DDC-AT~\cite{Xu2021b} shows minimal robustness,
while our models (CIRA3+AT-3 and PGD-AT-100) are
more robust. Perturbations over about 0.015 are already visible for the human eye.
For more details, please see \cref{sec:almaprox}.}
\label{fig:teaser}
\end{figure}

It has long been known that deep neural networks (and, in fact, most other machine learning
models as well) are sensitive to adversarial perturbation~\cite{Sturm2014a,szegedy13}.
In the case of image processing tasks, this means that---given a network and an input
image---an adversary can compute a specific input
perturbation that is invisible to the human eye yet changes the output arbitrarily.
This is not only a security problem but, more importantly, also a clue that the models
trained on image processing tasks have fundamental flaws regarding the feature representations
they evolve~\cite{adversarialbug19,engstrom19}.

In the context of image classification, this problem has received a lot of attention,
leading to a large number of attacks under various assumptions (just to mention a few,~\cite{carlini-wagner17,deepfool16,chen2020a,chen2020d,croce2022a}) and defenses (for example,~\cite{saddlepoint-iclr18,randsmoothing19}).
Currently, there is a generally accepted methodology for evaluating the robustness
of real-world networks~\cite{Carlini2019a}.
There is a leader-board as well, called RobustBench~\cite{Croce2021a}, where the models are evaluated
using AutoAttack~\cite{croce2020a}, a set of diverse and strong attacks.

In image segmentation, the area of robustness is less developed.
Obviously, there is a very strong connection with classification, since
classifier networks typically form the backbone of segmentation
networks~\cite{Zhao2017a,Chen2017a,Long2015a,Fu_2019_CVPR}.
This indicates that segmentation is equally vulnerable to adversarial perturbation.
Indeed, this vulnerability has been demonstrated many
times (for
example,~\cite{Cisse2017a,Arnab2018a,Xie2017a,Fischer2017a,Rony2022a,Sun2022a,Gupta2019a}).
Yet, these attacks are evaluated only on normally trained networks,
and only a few methods have been proposed for actually increasing robustness~\cite{Mao2020a,Klingner2020a,Xu2021b}.
To the best of our knowledge,
adversarial training---the only practically viable solution in image
classification---has been studied in depth only by Xu et al.~\cite{Xu2021b}.

Here, we propose an evaluation methodology more in line with the guidelines of
robustness research, in that we apply a set of different types of attacks with more aggressive
search parameters.
We demonstrate the utility of this evaluation methodology by showing that
the models in~\cite{Xu2021b} are not robust, and some simple adversarial training
strategies perform far better in this regard.


\subsection{Contributions}

\textbf{Thorough robustness evaluation.}
We demonstrate that it is essential to apply a combination of a set of diverse and strong
attacks to properly evaluate the robustness of segmentation models.
We propose the application of two types of attacks:
maximizing the error with a bounded perturbation,
and minimizing the perturbation for a given level of error.
Also, we use an aggressive hyperparameter setting for these attacks.
This way, we demonstrate that \emph{the networks created by Xu et al.~\cite{Xu2021b}
are not robust} but perform only gradient obfuscation~\cite{Athalye2018a}.

\textbf{Attacking the internal representations.} 
We propose novel attacks, that turn out to be crucial in revealing the lack of
robustness of the networks of Xu et al.~\cite{Xu2021b}.
Attacking the internal
representations to find adversarial perturbations is not novel in
itself in the context of segmentation~\cite{Sun2022a,Gupta2019a,Mopuri2019a}.
Unlike methods in related work,
we attack only a single layer, namely the final feature representation of the backbone
of the image segmentation network.
In addition, we also attack the output layer in the targeted version of the attack.
We also apply these attacks in adversarial training.

\textbf{Adversarial training.}
We train a number of robust models using the method of adversarial
training~\cite{saddlepoint-iclr18}.
To create adversarial examples during training, we experiment with two types of attacks:
the maximization of the cross-entropy loss, and decreasing the cosine similarity
of the internal feature representation.
The networks trained this way show nontrivial robustness according to our proposed
evaluation framework, outperforming the known baseline approaches by a large margin
(see \cref{fig:teaser} for a preliminary illustration).


\subsection{Related work}
\label{sec:related}

Here, we overview related work specifically in the area of adversarial
robustness in semantic segmentation.

\textbf{Attacks.}
Adaptations of the gradient-based adversarial attacks using the segmentation
loss function have been proposed relatively
early~\cite{Fischer2017a,Xie2017a,Arnab2018a}.
The Houdini attack uses a novel surrogate function more
tailored to adversarial example generation~\cite{Cisse2017a}.
Agnihotri et al. propose the cosine similarity as a surrogate~\cite{Agnihotri2023a}.
All these attacks work on the output of the segmentation network.
The ALMAProx attack proposed by Rony et al.~\cite{Rony2022a}
defines a constrained optimization problem to find the minimal perturbation to
change the prediction over a given proportion of pixels.
This is a fairly expensive, yet accurate baseline to evaluate defenses.

\textbf{Internal representation attacks.}
Similarly to the attack we propose here, there have been other proposals to target
the internal representations of the networks, as opposed to the output,
e.g.,~\cite{Sun2022a,Gupta2019a},
where several layers are attacked at the same time but these attacks are evaluated only on normally
trained networks.
Mopuri et al.~\cite{Mopuri2019a} 
also focus on internal activations as opposed to labels, but their goal is very different:
they wish to produce universal perturbations without looking at data.

\textbf{Special purpose attacks.}
Some works introduce different versions of the adversarial perturbation problem
with specific practical applications in mind.
Metzen et al. study universal (input independent) perturbations~\cite{Metzen2017a}.
Cai et al.~\cite{Cai2022a} study semantically consistent (context sensitive) attacks
that can fool defenses that look for semantic inconsistencies in the predicted scene.
Chen et al \cite{Chen2022b} also consider semantic attacks where
the predicted scene is still meaningful, only some elements are deleted, for example.

\textbf{Detection as a defense.} Klingner et al. proposed an approach to detect adversarial inputs based on the consistency on
different tasks~\cite{Klingner2022a}.
Another detection approach was suggested by Bär et al.~\cite{Baer2019a,Baer2020a}
where a specifically designed ensemble of models is applied that involves
a dynamic student network that explicitly attempts to become different from
another model of the ensemble.
While detection approaches
are useful in practice, they are not hard defenses because adversarial inputs
can be constructed to mislead multiple tasks or multiple models
as well simultaneously, as the authors also note.

\textbf{Multi-tasking as a defense.}
Another approach involves multi-task networks, with the underlying idea that
if a network is trained on several different tasks then it will naturally become
more robust~\cite{Mao2020a,Klingner2020a}.
While this is certainly a very promising direction for defense, many uncertainties
are involved such as generalizability to different tasks and datasets, and the critical
number of tasks.

\textbf{Adversarial training.}
In image classification,
the most successful approach is adversarial training~\cite{saddlepoint-iclr18}.
In semantic segmentation, a notable application of adversarial training
is the DDC-AT algorithm by Xu et al.~\cite{Xu2021b}.
This approach will serve as one of our test cases that we evaluate using our
evaluation methodology.
More recently, the SegPGD\footnote{%
The implementation of SegPGD is not yet publicly available, but we will extend our evaluations
when it becomes available, and we will update this manuscript.}
algorithm was proposed by Gu et al.~\cite{Gu2022a}, an attack
that was also used to implement adversarial training.

\section{Background}
\label{sec:back}

Here, we discuss the basic notions of adversarial attacks and adversarial training in
the white-box setting.

\subsection{Adversarial Attacks}

We assume that we are given a pre-trained model
$f_\theta: \mathcal X\mapsto \mathcal Y$ and a
loss function $\mathcal L(\theta, x, y)\in\mathbb R$ that characterizes the error of the
prediction $f_\theta(x)$ given the ground truth output $y\in \mathcal Y$.

Intuitively, the goal of the adversarial attack is to find a very small perturbation of a
given input $x$ in such a way that the prediction of the model $f_\theta(x)$ is completely
wrong.
A possible way of formalizing this vague objective is the maximization problem
\begin{equation}
\delta^*=\arg\max_{\delta\in\Delta} \mathcal L(\theta, x+\delta, y),
\label{eq:untargeted-obj}
\end{equation}
which gives us the perturbed input $x+\delta^*$ that causes the most damage in terms
of the loss function.
Here, the set $\Delta$ captures the idea of ``very small perturbation''.

In this paper, we will work with $\mathcal X=[0,1]^{H\times W\times 3}$, that is,
the inputs are 3-channel color images of width $W$ and height $H$, with all the
values normalized into the interval $[0,1]$.
We adopt the widely used definition $\Delta=\{\delta: \|\delta\|_\infty\leq\epsilon\}$
that defines the neighborhood of an input $x$ in terms of the maximal absolute difference
in any tensor value.

\textbf{FGSM.} Many algorithms have been proposed for solving or approximating this maximization problem.
A very efficient but also rather inaccurate approximation is the fast gradient sign method
(FGSM)~\cite{goodfellow15} that uses the approximation
\begin{equation}
\delta^*\approx \epsilon\,\mathrm{sgn}\, \nabla_x \mathcal L(\theta, x, y).
\label{eq:fgsm}
\end{equation}

\textbf{PGD.} The iterative version of the same approach, usually referred to as the
projected gradient descent (PGD) attack~\cite{Kurakin2017a,saddlepoint-iclr18} is a widely used
attack today due to its simplicity and efficiency.
The attack starts from $x_0=x$ and computes the iteration
\begin{equation}
x_{t+1}=
\mathrm{Proj}_{x, \epsilon} 
(x_t+\alpha\,\mathrm{sgn}\, \nabla_{x_t} \mathcal L(\theta, x_t, y)),
\label{eq:pgd}
\end{equation}
where Proj$_{x,\epsilon}(z)$ clips each coordinate $z_i$ into the interval
$[x_i-\epsilon,x_i+\epsilon]$.
Apart from $\epsilon$, the PGD attack has two additional important hyperparameters: the number
of iterations, and the step size $\alpha$.

\textbf{Targeted attacks.}
So far, the goal was simply to move away from the ground truth prediction as much
as possible.
However, in some applications we might want to force the model to produce a
specific incorrect output $y'$.
This problem can be formulated as the minimization problem
\begin{equation}
\delta^*=\arg\min_{\delta\in\Delta} \mathcal L(\theta, x+\delta, y'),
\end{equation}
which is very similar to the untargeted maximization problem except we replace
$y$ with $y'$, and minimize.
The FGSM and PGD algorithms can trivially be adopted to the targeted setting.

\subsection{Adversarial Training}

So far, the most reliable heuristic solution to achieve robustness is
through adversarial training~\cite{goodfellow15,saddlepoint-iclr18}.
The idea behind adversarial training is to use adversarial examples during
training as an augmentation.
The adversarial examples are always created based on the current model in the given
update step.
Formally, we wish to solve the following learning task:
\begin{equation}
\label{eq:at-task}
    \theta^*=\arg\min_\theta \mathbb{E}_{p(x,y)}[
    \max_{\delta \in \Delta} \mathcal{L}(x + \delta), y)],
\end{equation}
where $p(x,y)$ is the distribution of the data.

In the outer minimization (learning) task one can use an arbitrary learning method, and
in the inner maximization task one can select any suitable attack to
perturb the samples used by the learning algorithm.

\section{Cosine Internal Representation Attacks}

In this section, we discuss our variants of internal representation attacks
against segmentation networks.
We will use these attacks as part of our attack collection for evaluation, as well as in
an adversarial training algorithm.

\subsection{Attacking Segmentation Networks}

In the case of the semantic segmentation task, the methods described in
\cref{sec:back} are all applicable~\cite{Fischer2017a,Xie2017a,Arnab2018a}.
In this case, the output space $\mathcal Y$ is $[0,1]^{H\times W\times C}$, where
$C$ is the number of possible categories for each pixel.
Note that $\mathcal Y$ is the softmax output, which defines a probability distribution for
each pixel over the categories that can be used to compute the
final segmentation mask by taking the maximum probability category.
(Recall, that the input was defined in \cref{sec:back} as
$\mathcal X=[0,1]^{H\times W\times 3}$, that is,
the inputs are 3-channel color images of width $W$ and height $H$). 

Throughout the paper, the loss function will be assumed to be the cross entropy
loss, given by
\begin{equation}
\mathcal L(\theta, x, y) = 
\frac{1}{H W}\sum_{h=1}^H\sum_{w=1}^W\sum_{c=1}^C -y_{h,w,c}\log f_\theta(x)_{h,w,c}.
\end{equation}

Apart from the standard attacks we have described, we will now discuss attacks targeting
the internal representation of segmentation networks.

\subsection{Attacking the Internal Representations}

As we discussed in \cref{sec:related}, some attacks in previous work are based
on targeting the internal layers of networks.
Here, we will introduce our own variant of such an attack with adversarial training in mind,
which requires the attacks to be simple and efficient.

First, we need to represent the model $f_\theta$ in finer resolution.
In this study,
we focus on architectures that are based on a backbone network that is extended with
a segmentation head (for example, DeepLabv3~\cite{Chen2017a} and PSPNet~\cite{Zhao2017a}).
Let $f^b_{\theta_b}$ represent the backbone network, and $f^h_{\theta_h}$ the head.
In this framework, we have $f_\theta(x)=f^h(x,f^b_{\theta_b}(x))$.

Our first attack is based on targeting only the output of the backbone.
In our preliminary study, we found that the most effective approach
is to maximize the cosine similarity of the feature representation
of the perturbed input and a random vector $z$, where each element $z_i$ is
sampled uniformly at random from the interval $[0,1]$.
Note that the length of $z$ is irrelevant, because cosine similarity, given by
\begin{equation}
\mathrm{CosSim}(x,y) = \frac{x\cdot y}{\|x\|_2\|y\|_2}
\end{equation}
is insensitive to the lengths of the vectors, as it focuses on the angle between them.
Formally, we wish to solve the maximization problem
\begin{equation}
\delta^*=\arg\max_{\delta\in\Delta} \mathrm{CosSim}(f^b_{\theta_b}(x+\delta), z ).
\label{eq:cira}
\end{equation}

It might seem counter-intuitive at first to use a random vector $z$ as a target
for our attack.
However, a random vector has a very low cosine similarity from any vector with a
high probability, due to the high dimensionality of the internal
representation~\cite{Blum2020a}.

\textbf{CIRA.}
To solve the maximization problem in \cref{eq:cira} we use the Adam optimizer~\cite{adam15}
combined with projection, given that we are working with a constrained problem.
From now on, we will refer to this attack as CIRA (cosine internal
representation attack).

Applying Adam in this context is straightforward, it involves only one simple
modification relative to the unconstrained application of Adam:
after every Adam update step, we clip the current vector $x_t$
into the set $\Delta$, as was done in the PGD algorithm in \cref{eq:pgd}.

\textbf{Targeted CIRA.}
Since the internal representation is not directly computable from the ground truth label
$y$, implementing a targeted version is challenging.
A possible approach is to replace the target ground truth $y'$ with the internal
representation of an input image $x'$ that has $y'$ as ground truth.
Note that this does not allow us to use arbitrary targets, only those that
have a corresponding input image.
The attack involves solving the maximization problem
\begin{equation}
\delta^*=\arg\max_{\delta\in\Delta} \mathrm{CosSim}(f^b_{\theta_b}(x+\delta), f^b_{\theta_b}(x') ),
\label{eq:targeted-cira}
\end{equation}
which is the same problem as \cref{eq:cira} but with $z$ replaced with $f^b_{\theta_b}(x')$.
As before, we use projected Adam to solve this problem.

\subsection{CIRA+: a Hybrid Attack}
We introduce a hybrid attack as well, that we coined CIRA+.
This attack targets the internal representation
as well as the softmax output simultaneously.
The motivation is to utilize the power of the internal attack and the precision of the
softmax layer attack, especially in the case of the targeted scenario.
The maximization problem corresponding to the untargeted version of CIRA+ is 
\begin{equation}
\begin{split}
\delta^* = & \arg\max_{\delta\in\Delta} \left[ \vphantom{f^b_q} \alpha\mathcal L(\theta, x+\delta, y)+\right.\\
& \left. (1-\alpha)\mathrm{CosSim}(f^b_{\theta_b}(x+\delta), z )\right],
\end{split}
\label{eq:cirap}
\end{equation}
where $\alpha\in[0,1]$ and $z$ is a random vector.
The objective is the weighted average of the softmax loss objective
in \cref{eq:untargeted-obj} and the objective in \cref{eq:cira}.
As in previous cases, we solve this optimization problem using projected Adam.
During our experiments, we fixed $\alpha=0.5$.

\textbf{Targeted CIRA+.}
The targeted version of the attack objective is also the hybrid of the
softmax and internal objectives,
and thus inherits the constraint that the target mask $y'$ has to belong to an
existing input example $x'$.
But in this case, we will use $y'$ both directly and indirectly, according to the
targeted objective
\begin{equation}
\begin{split}
\delta^* = & \arg\min_{\delta\in\Delta} \left[ \vphantom{f^b_q} \alpha\mathcal L(\theta, x+\delta, y')-\right.\\
& \left. (1-\alpha)\mathrm{CosSim}(f^b_{\theta_b}(x+\delta), f^b_{\theta_b}(x') )\right].
\end{split}
\label{eq:targeted-cirap}
\end{equation}
where $\alpha\in[0,1]$.
Here, the term involving the cosine similarity is negated because similarity has to be
maximized to the target.
As before, this problem is solved using projected Adam.

\textbf{PAdam.}
For completeness, we include yet another attack in our list, that we coined PAdam.
This attack is very similar to PGD, in that it solves the same objective, namely
\cref{eq:untargeted-obj}, but instead of using the signed gradient steps, we use
projected Adam.
This attack can also be thought of as a special case of CIRA+ with $\alpha=1$.

\subsection{CIRA vs. CIRA+}


In terms of computational cost, CIRA is a very efficient attack because
it uses only the backbone network $f_{\theta_b}^b$, unlike FGSM or PGD.
In terms of untargeted attack success, it is comparable to traditional options
like PGD, as we will see later.

On the other hand, CIRA+ is more expensive because it works on the entire network.
However, on some networks its attack success is vastly superior to both CIRA and PGD,
as our evaluation results will show.


\begin{table}
\caption{Summary of investigated models.}
\vspace{2mm}
\label{table:models}
\centering
\setlength{\tabcolsep}{5pt}
\begin{tabular}{l|p{6cm}}
Name & Attack used in adversarial training\\\hline
CIRA+AT-2 & CIRA+ with 2 iterations, Adam learning rate: $0.01$.\\
CIRA+AT-3 & CIRA+ with 3 iterations, Adam learning rate: $0.01$.\\
PGD-AT-50 & PGD, step size: $0.01$, iterations: 3, 50\% adversarial batch \\
PGD-AT-100 & PGD, step size: $0.01$, iterations: 3, 100\% adversarial batch \\
DDC-AT & Model Instance used in~\cite{Xu2021a}\\
Normal & None\\
\end{tabular}
\end{table}

\begin{table*}[tb]
\caption{mIoU values. Iteration number is indicated in parenthesis.}
\vspace{2mm}
\label{table:pspnet-bounded}
\centering
\setlength{\tabcolsep}{5pt}
\begin{tabular}{l|cccccc|c}
Model      &  No attack  &  PGD(6)     & PGD(120)    & PAdam(120)  & CIRA(120)   & CIRA+(120) & MIN\\\hline
PGD-AT-100 & 42.47       & 19.01       & 11.32       & 11.19       & {\bf 35.09} & {\bf 13.79} & {\bf 10.13} \\
CIRA+AT-3  & 54.88       & 18.41       & 11.04       & 10.02       & 25.66       & 12.27       & 9.60 \\
CIRA+AT-2  & 57.53       & 17.53       & 6.43        & 6.27        & 19.41       & 9.91        & 5.15 \\ 
PGD-AT-50  & 69.00       & 25.80       & {\bf 14.66} & {\bf 14.83} & 14.38       & 0.72        & 0.68 \\
DDC-AT     & 71.70 & {\bf 28.70} & 14.28 & 12.96 & 10.89 & 0.53 & 0.51 \\
Normal  & {\bf 74.6} & 2.1 & 0.64 & 0.45 & 0.99 & 0.42 & 0.36 \\
\end{tabular}
\end{table*}

\section{Investigated Models}

Here, we describe the model instances included in our empirical investigation, including
our models and the baseline models.
Every model we investigated was trained on the Cityscapes dataset~\cite{Cordts2016a}
with the PSPNet~\cite{Zhao2017a} architecture.

To evaluate our proposed adversarial training algorithm, we trained a number of
robust models using our method to compare the robustness of these models
with related work using our proposed evaluation methodology.

The training of every model in our investigation shared the
hyperparameters that are recommended in the semseg package~\cite{semseg2019}.
These settings were also used by Xu et al.~\cite{Xu2021b}.
In more detail, the batch size was 16, we used SGD with momentum as the learning algorithm
with a dynamic learning rate (starting at $0.01$),
a weight decay coefficient of $10^{-4}$, and a number of augmentations including rotation,
scaling and cropping.
For more details, please refer to our implementation.

The internal attacks applied in adversarial training used the $\ell_\infty$-norm neighborhood
$\Delta=\{\delta: \|\delta\|_\infty\leq\epsilon\}$ with $\epsilon=0.03$,
in line with the general practice.
Note that this assumes that the input channels are scaled to the range $[0,1]$.

\textbf{Reference models.}
We used three reference models.
The first one is a normally trained model.
The next two models are the pre-trained models available from
the implementation repository of~\cite{Xu2021a}.
One of these was adversarially trained using an internal PGD attack with three 
iterations.
The other one is the divide-and-concur model proposed in~\cite{Xu2021a}.
It is important to note that
these two models were trained on batches of 50\% normal and 50\% adversarial examples.

\textbf{CIRA+ models.} 
We used the CIRA+ attack in adversarial training
with two different settings: two or three iterations.
We will refer to these model instances as CIRA+AT-2 and CIRA+AT-3.
For these models we used early stopping based on the performance of
a 10-epoch window over the validation set. 
The models we used were from epochs 1103 and 469, respectively.
All of our models used only adversarial examples for training,
as opposed to our reference models.
We used the default parameters for Adam, with a learning rate of $0.01$.

\textbf{PGD.}
We also trained an additional model using a 3-iteration PGD, but this time
using 100\% adversarial examples, as opposed to only 50\%.
In this training we also used early stopping, and our model instance is
from epoch 487.

\Cref{table:models} contains a summary of the investigated models.

\section{Robustness to Bounded Attacks}
\label{sec:exp-bounded}


Here, we investigate the robustness of our set of models to an array of attacks
that generate bounded perturbations from
$\Delta=\{\delta: \|\delta\|_\infty\leq\epsilon\}$ with $\epsilon=0.03$, assuming
the input values are in $[0,1]$.
We used the attacks described previously, with the parameters that were used also
in the adversarial training algorithm, except for the iteration number that we will
indicate in each case.

The results are shown in \cref{table:pspnet-bounded}.
The table contains the mean of class-wise intersection-over-union (mIoU) scores
for each model and attack, evaluated over the test set of the Cityscapes dataset.
Note that the last column indicates an aggregated score, which is computed by
taking the minimum score of the set of attacks for each individual test example, and
averaging over these minimum scores.
This aggregation methodology is very similar to the one used in AutoAttack~\cite{croce2020a}.

\begin{figure}
\centering
\setlength{\tabcolsep}{1pt}
\begin{tabular}{cc}
Clean Image & Ground Truth Label\\
\includegraphics[width=0.48\columnwidth]{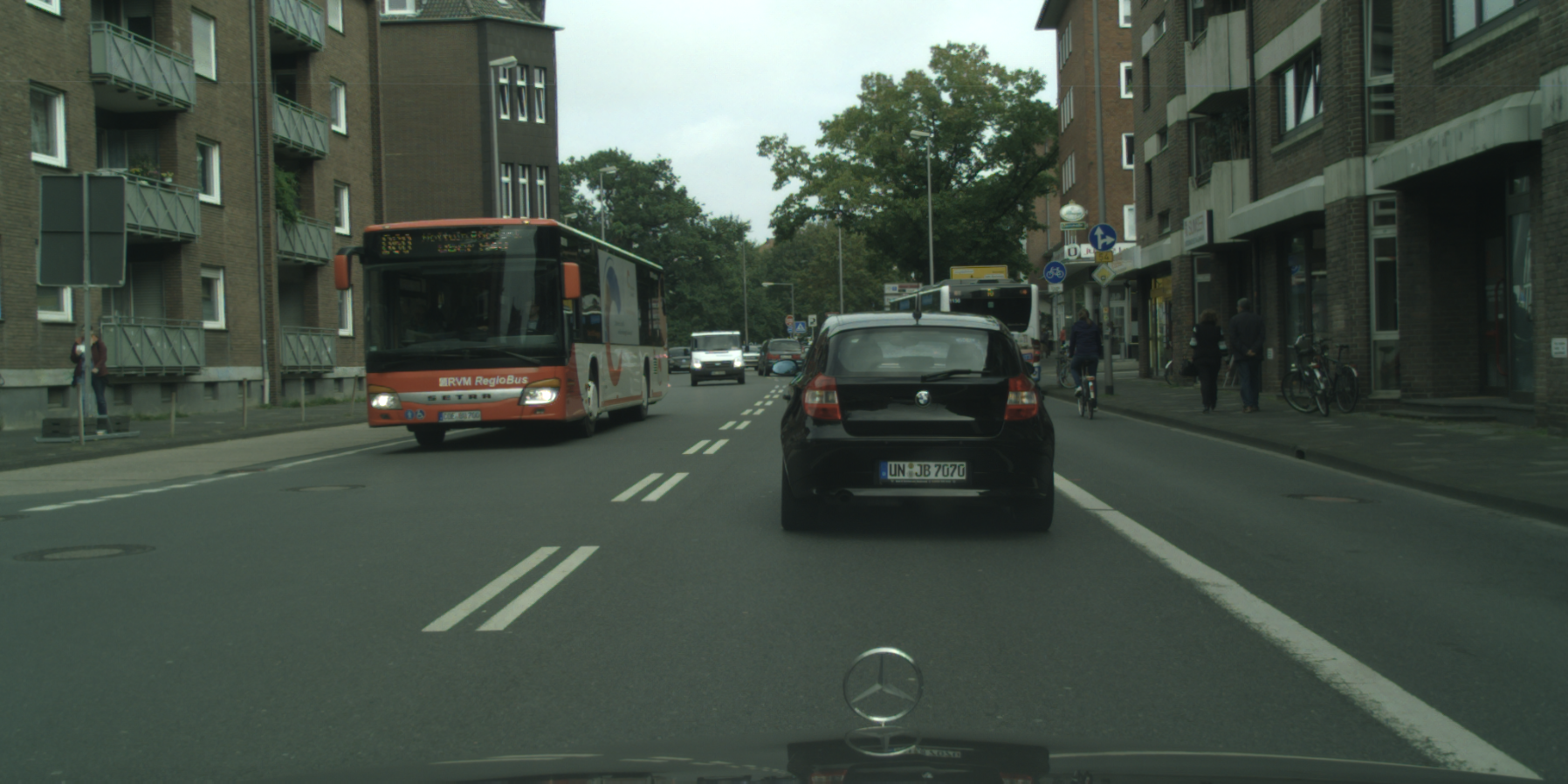}&
\includegraphics[width=0.48\columnwidth]{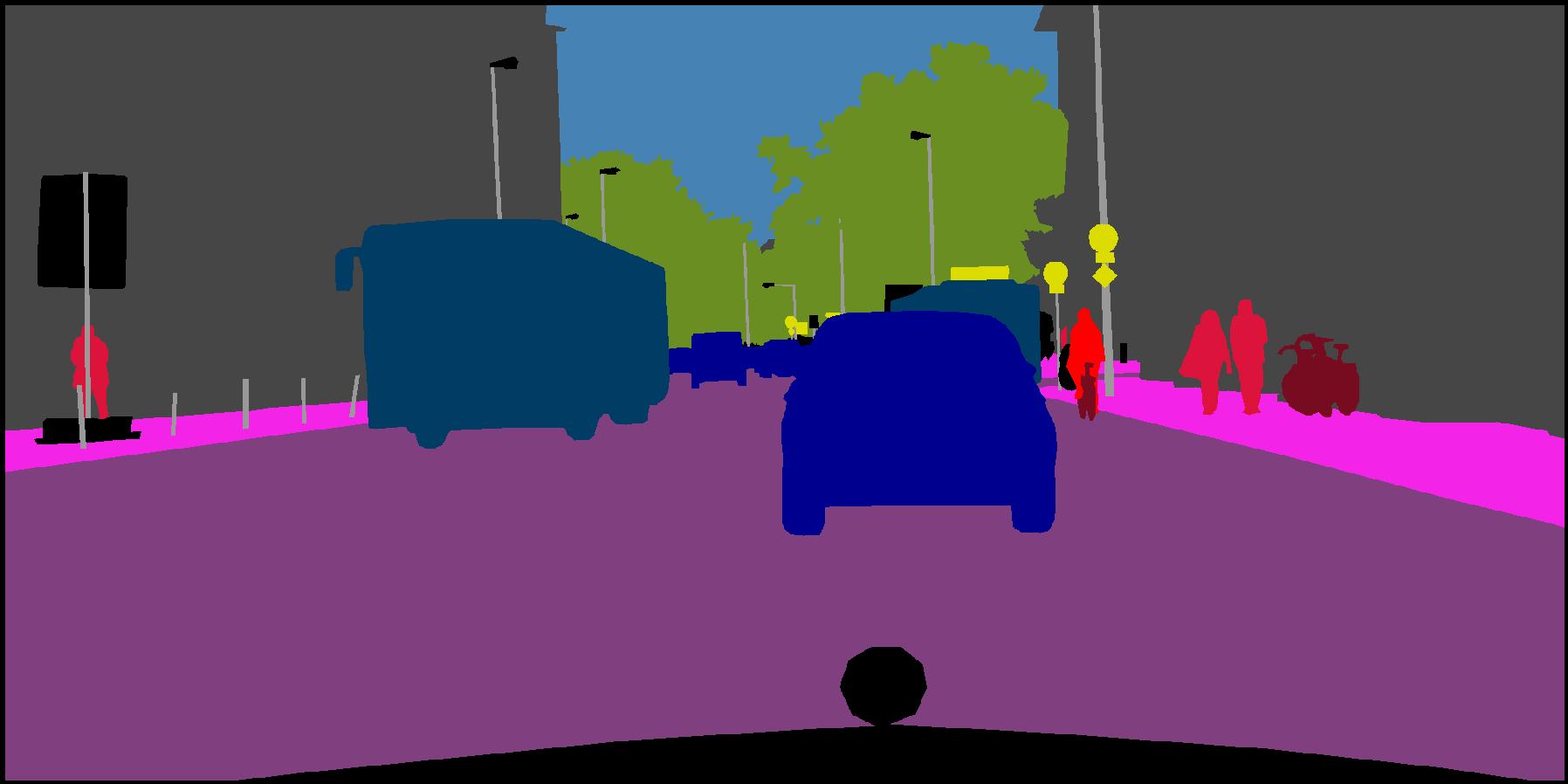}\\
Perturbed Image & Predicted Label\\
\includegraphics[width=0.48\columnwidth]{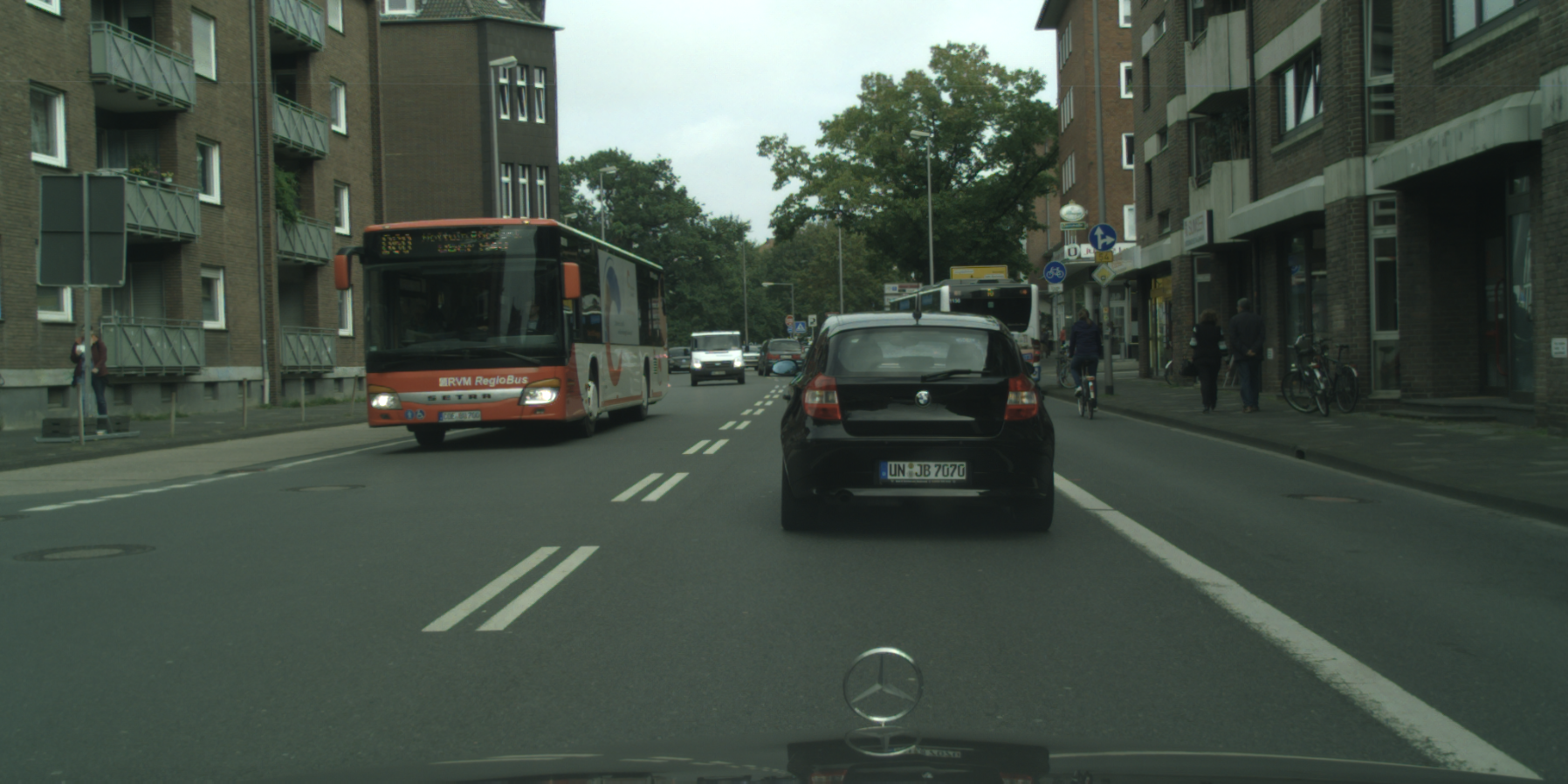}&
\includegraphics[width=0.48\columnwidth]{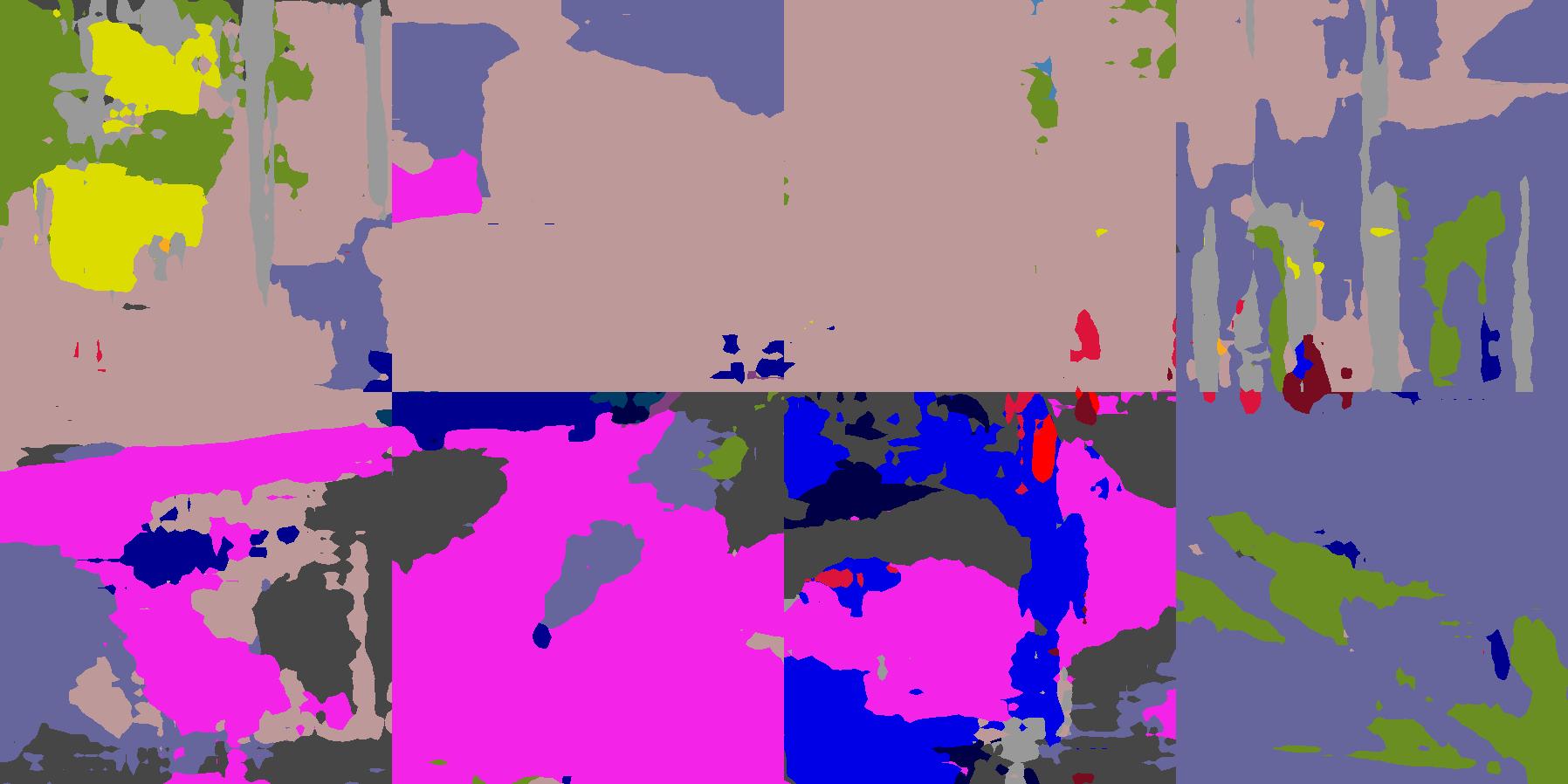}\\
\end{tabular}
\caption{Visual illustration of the vulnerability of the Normal network.
The sample image was selected at random.
We applied the ALMAProx attack  (see \cref{sec:almaprox}) to get a very small,
invisible perturbation of $\ell_\infty$ norm 0.0011 with a pixel error of 99.18\%.}
\label{fig:examples2}
\end{figure}

\begin{figure}
\centering
\setlength{\tabcolsep}{1pt}
\begin{tabular}{ccc}
& Perturbed Image & Predicted Label\\
\multicolumn{3}{c}{CIRA+AT-3}\\
\rotatebox{90}{\parbox{2cm}{\centering CIRA(120)}}&
\includegraphics[width=0.48\columnwidth]{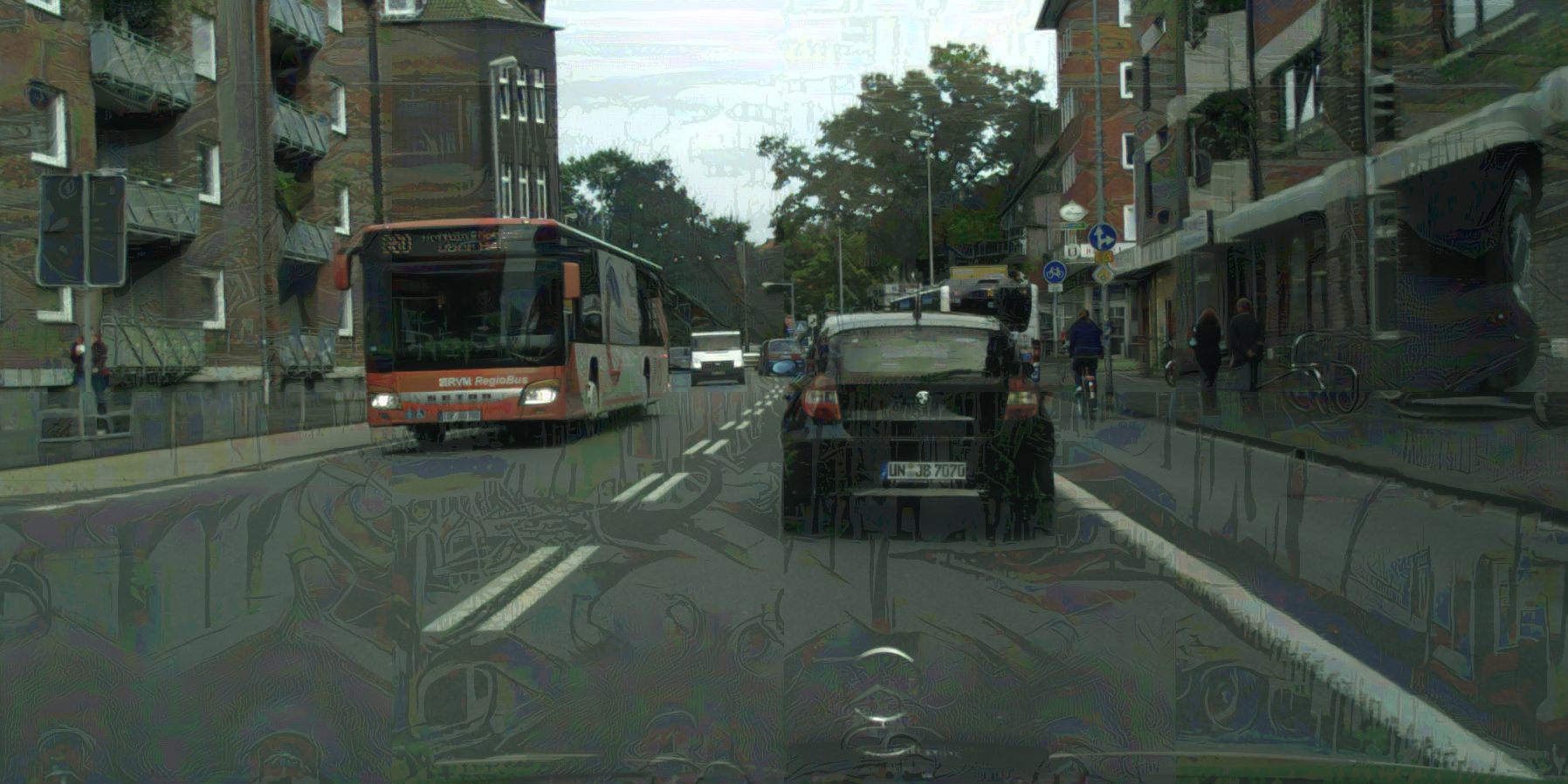}&
\includegraphics[width=0.48\columnwidth]{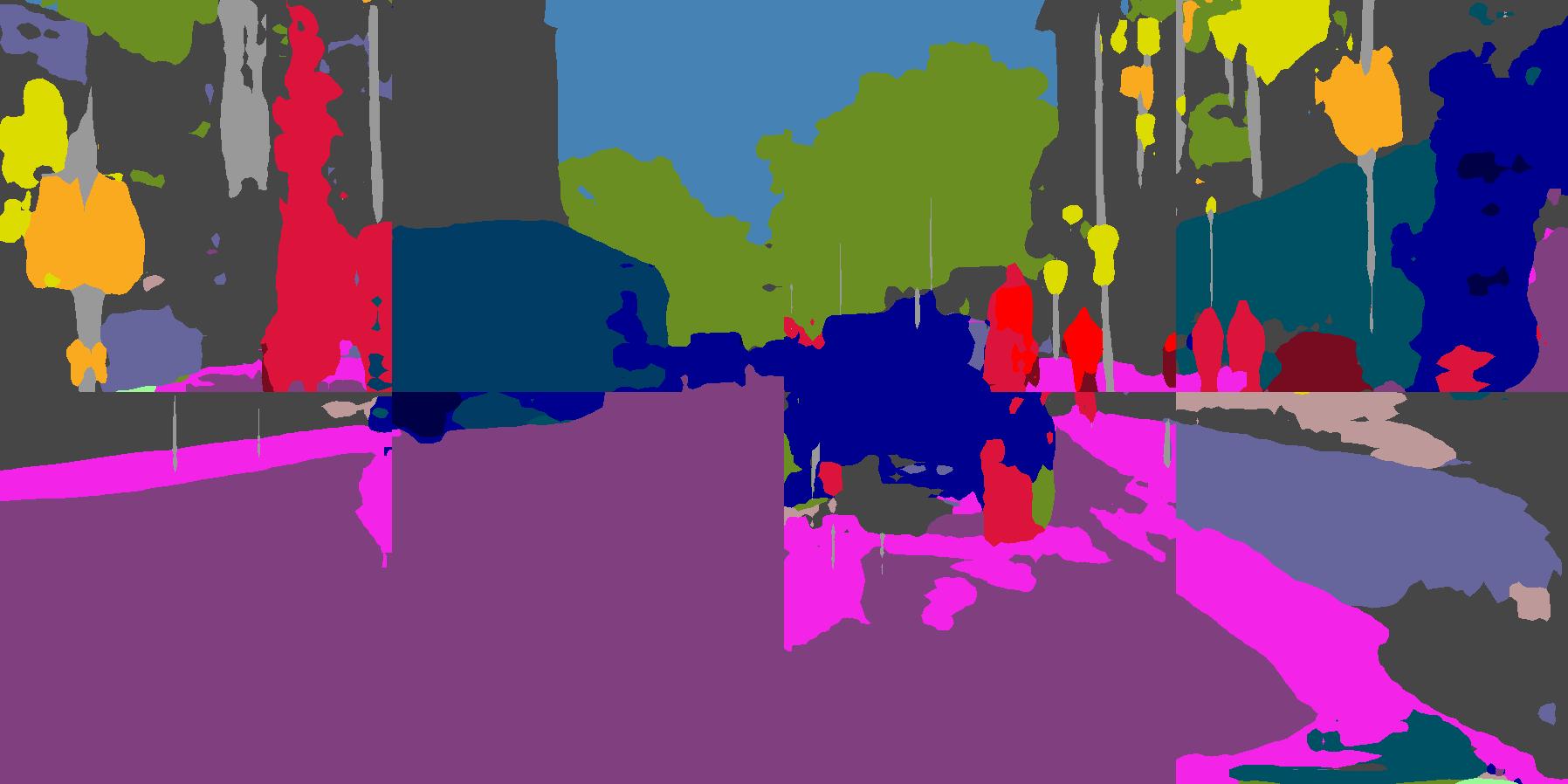}\\
\rotatebox{90}{\parbox{2cm}{\centering PGD(120)}}&
\includegraphics[width=0.48\columnwidth]{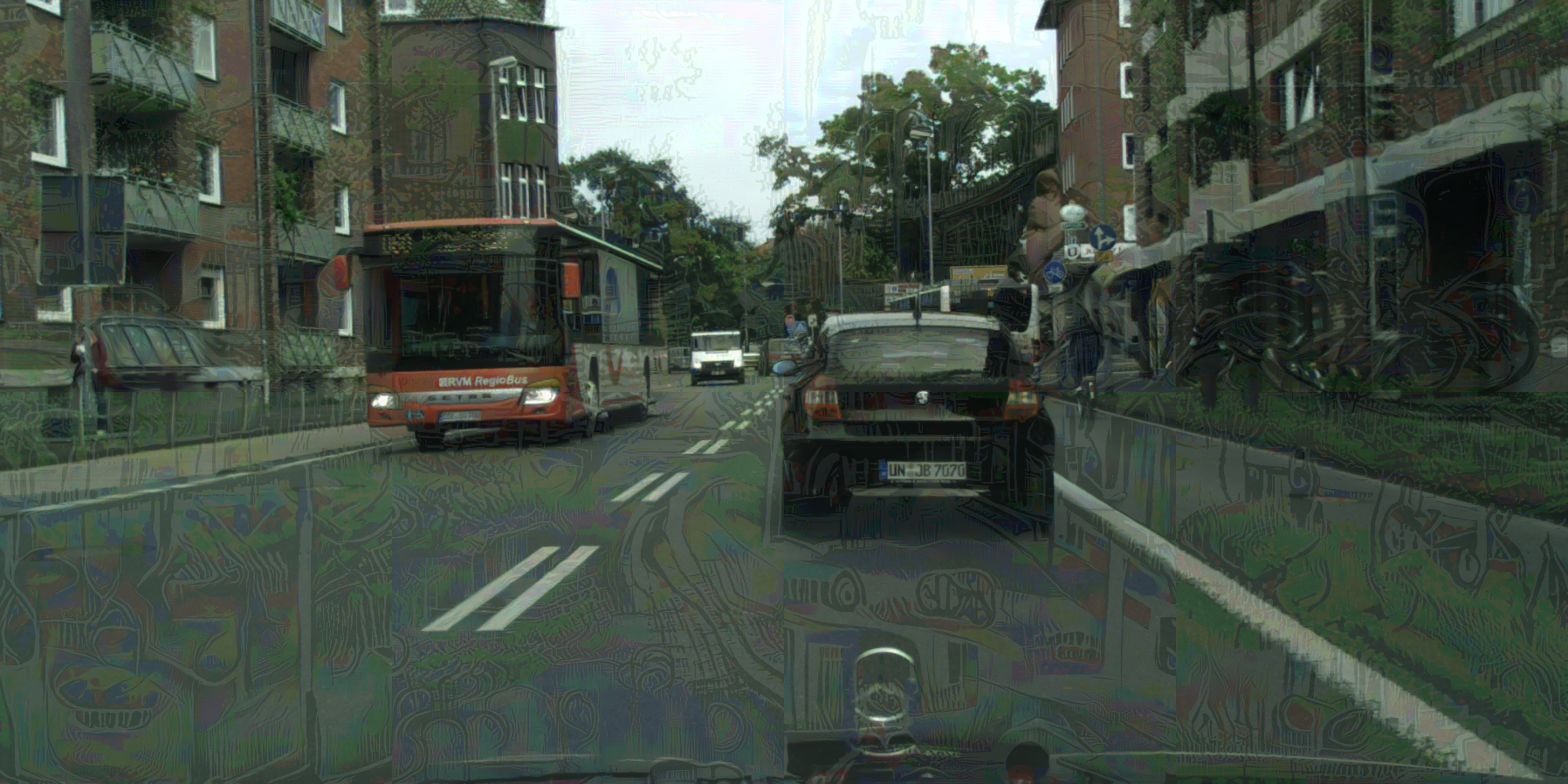}&
\includegraphics[width=0.48\columnwidth]{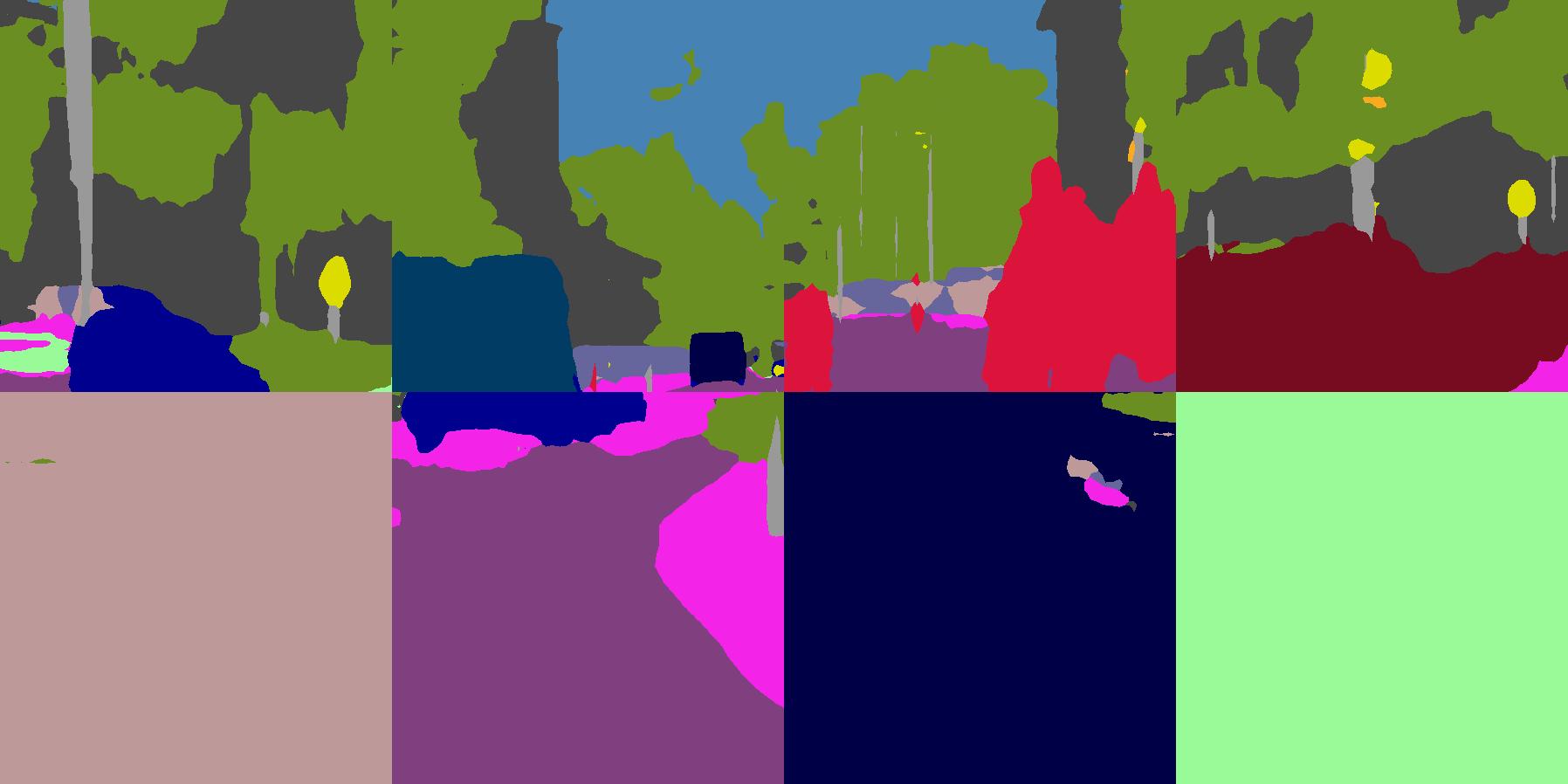}\\
\rotatebox{90}{\parbox{2cm}{\centering CIRA+(120)}}&
\includegraphics[width=0.48\columnwidth]{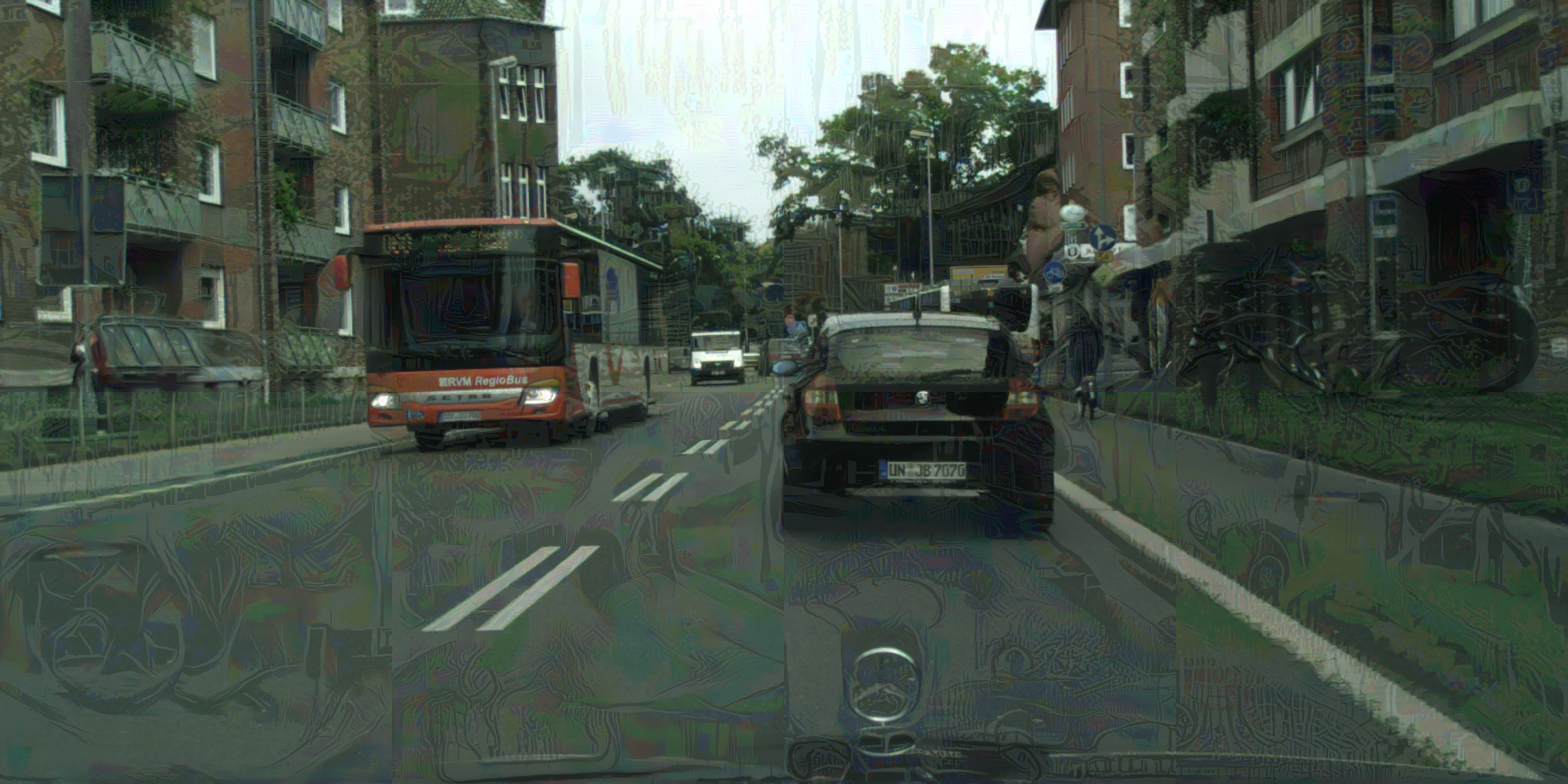}&
\includegraphics[width=0.48\columnwidth]{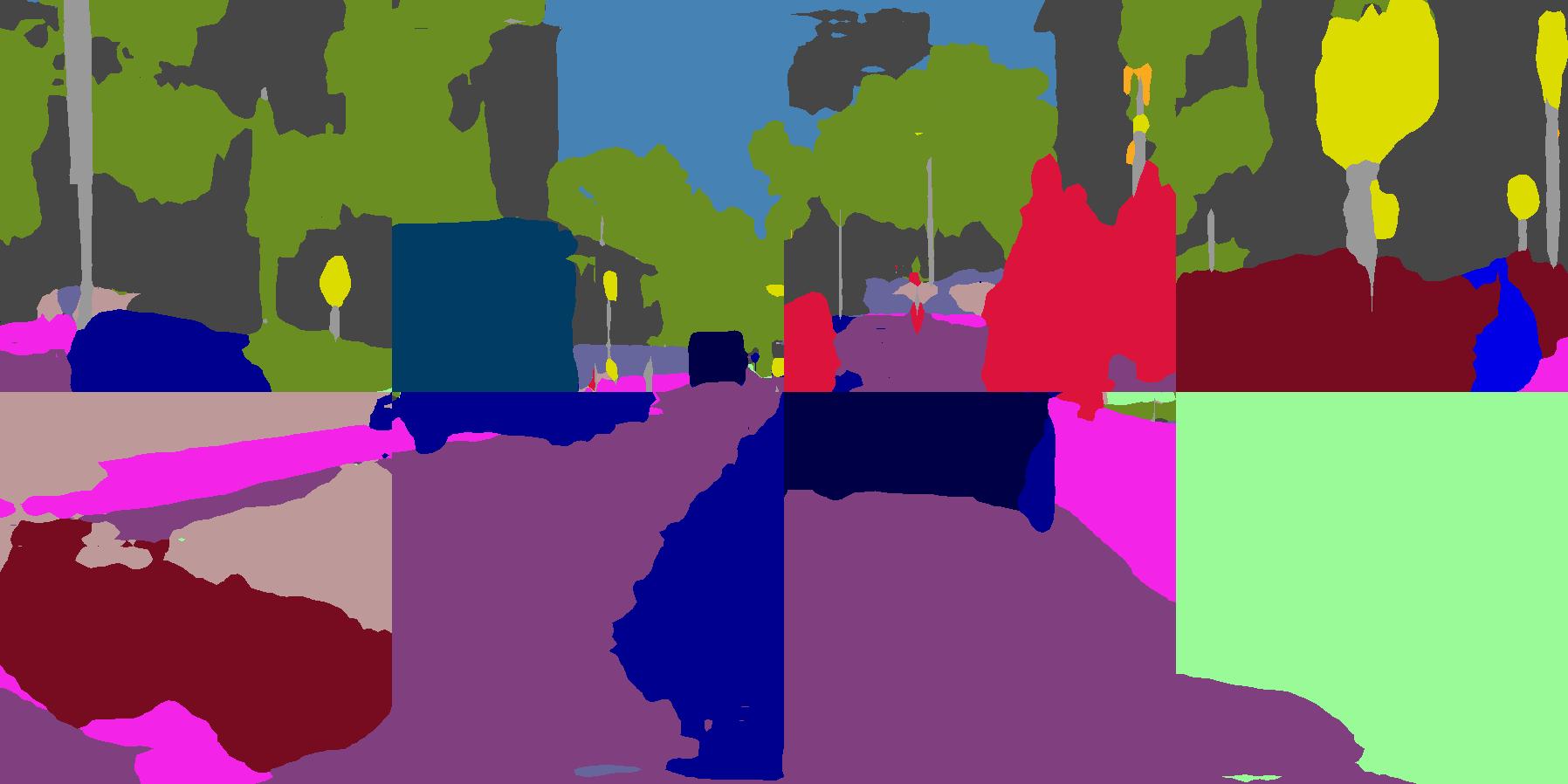}\\
\multicolumn{3}{c}{PGD-AT-100}\\
\rotatebox{90}{\parbox{2cm}{\centering CIRA(120)}}&
\includegraphics[width=0.48\columnwidth]{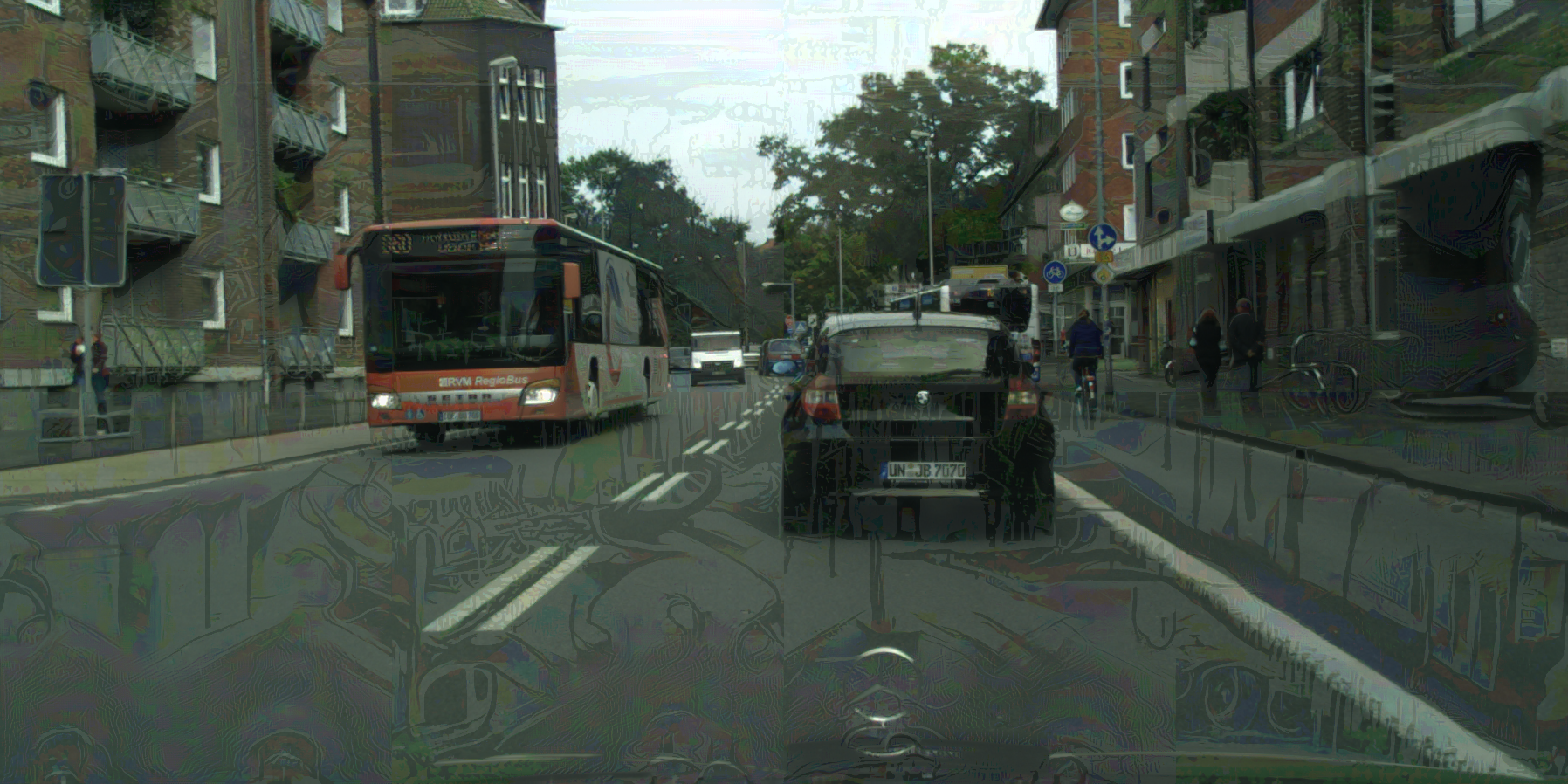}&
\includegraphics[width=0.48\columnwidth]{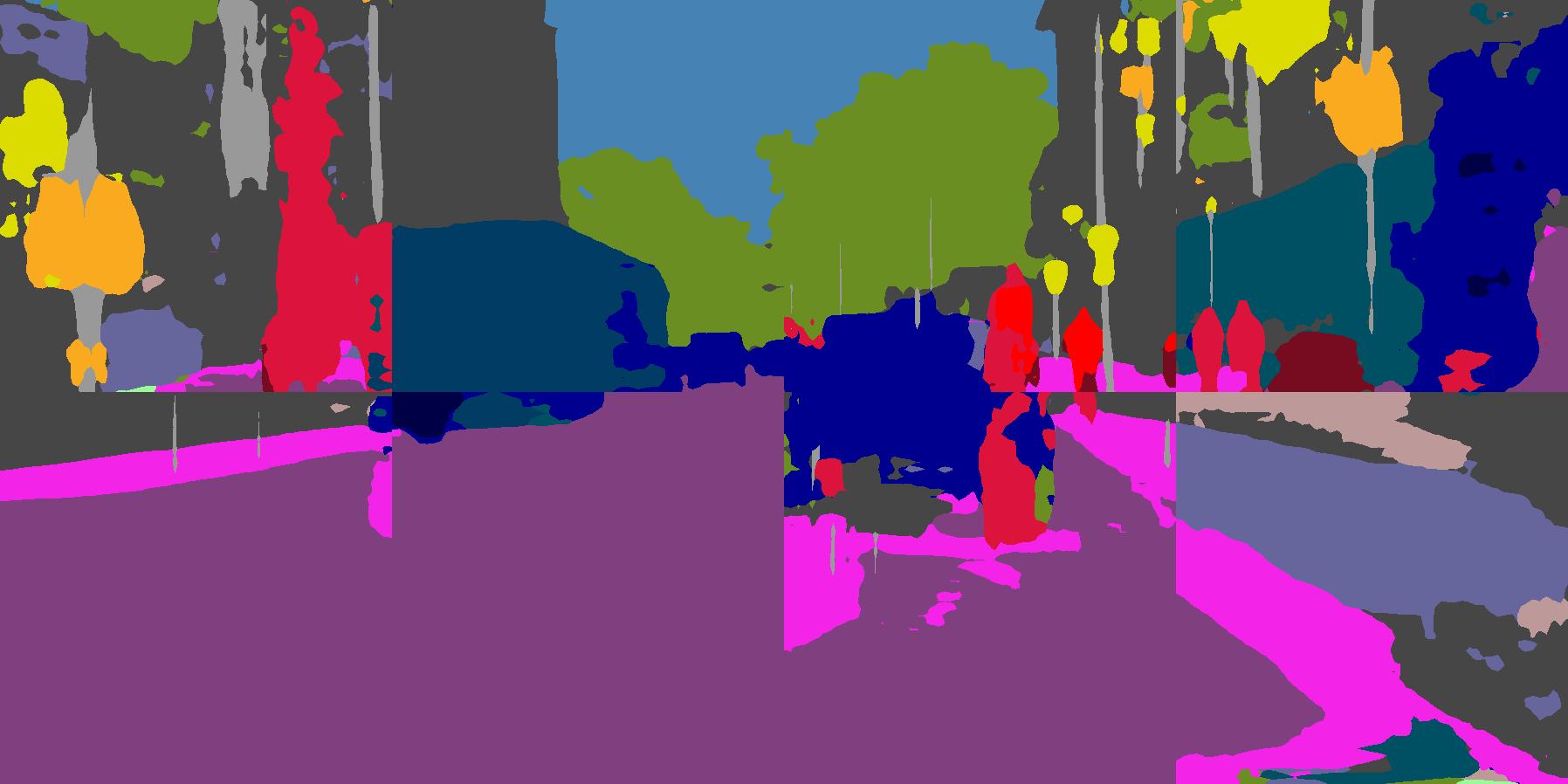}\\
\rotatebox{90}{\parbox{2cm}{\centering PGD(120)}}&
\includegraphics[width=0.48\columnwidth]{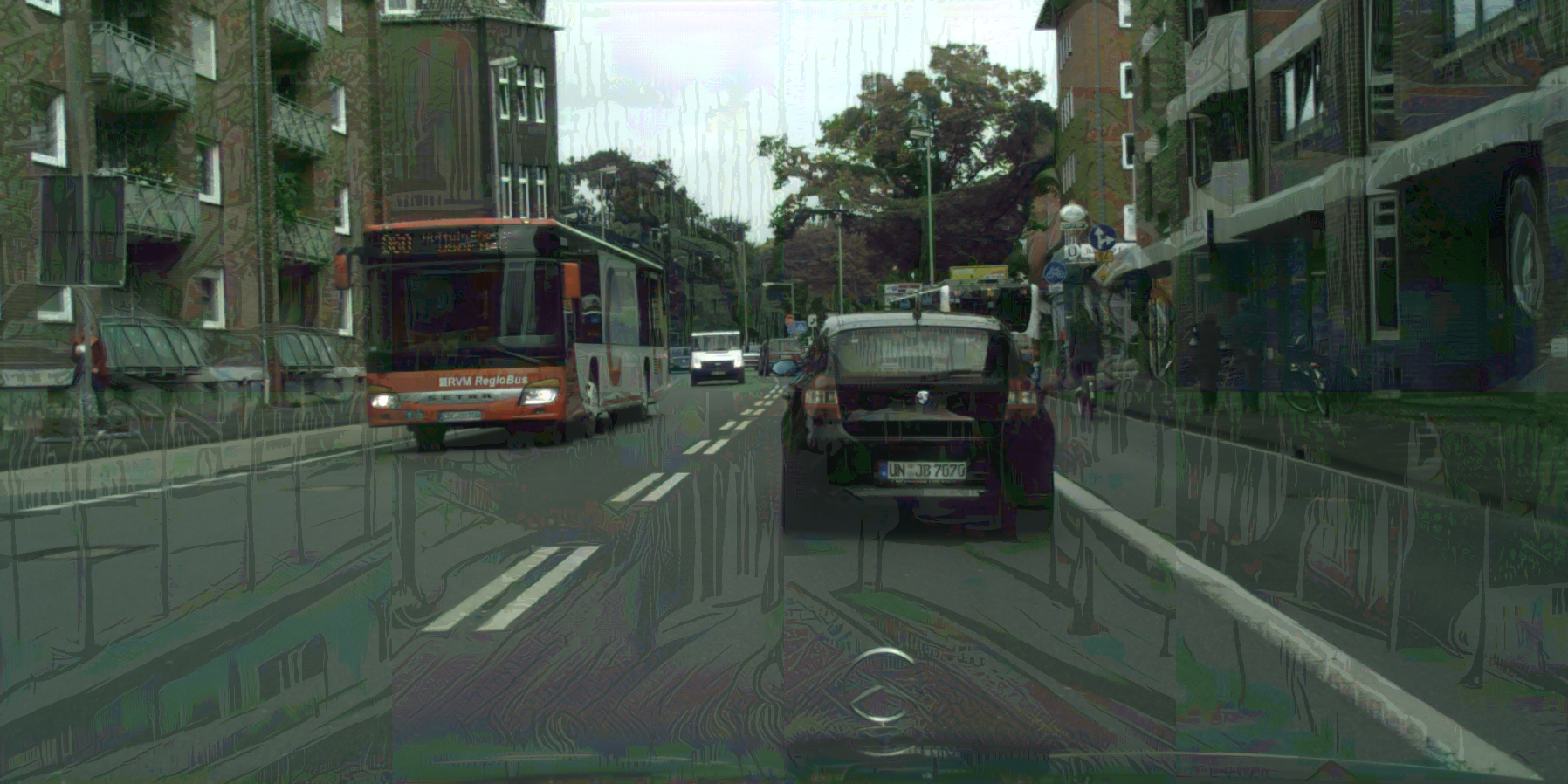}&
\includegraphics[width=0.48\columnwidth]{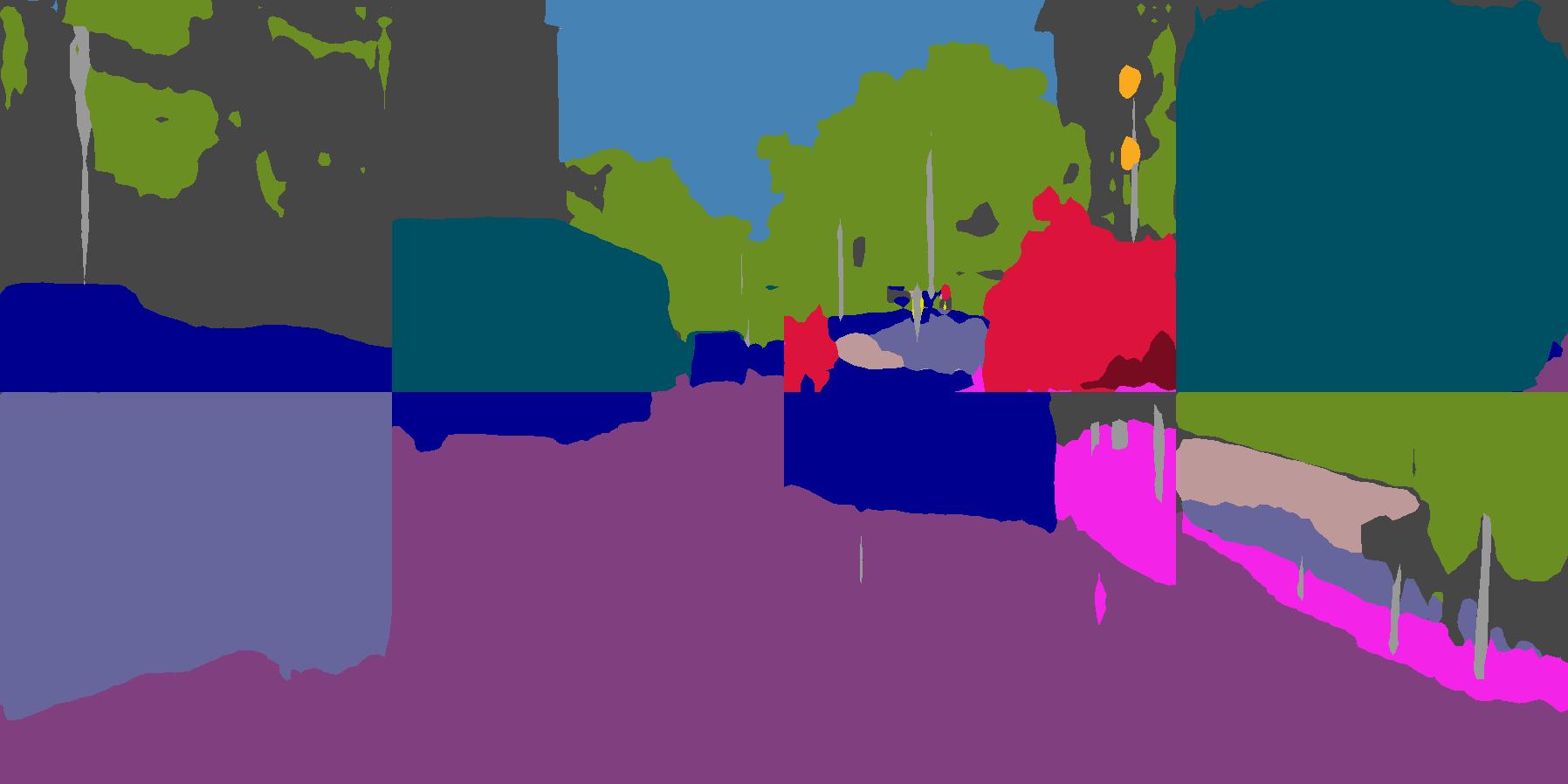}\\
\rotatebox{90}{\parbox{2cm}{\centering CIRA+(120)}}&
\includegraphics[width=0.48\columnwidth]{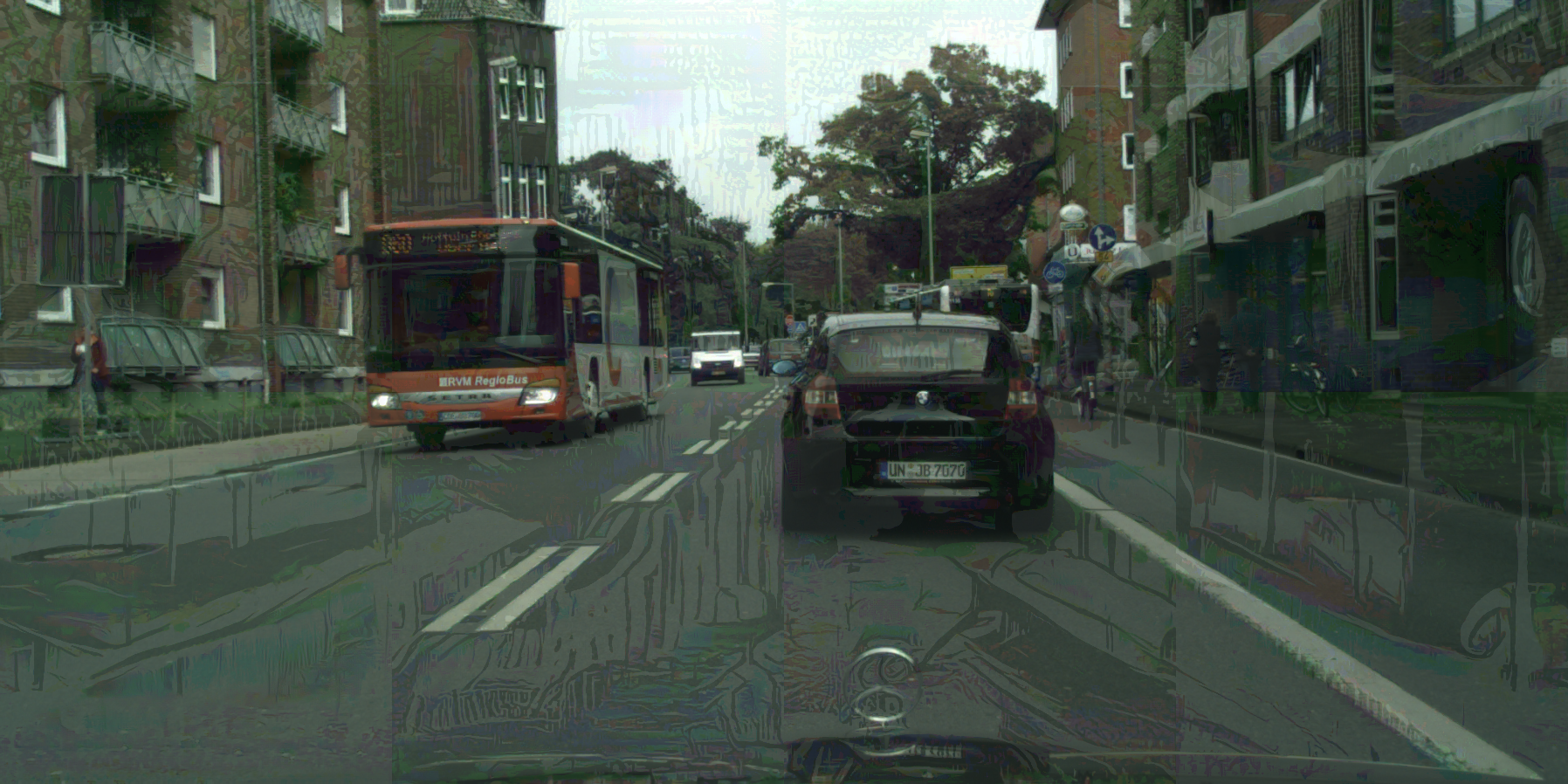}&
\includegraphics[width=0.48\columnwidth]{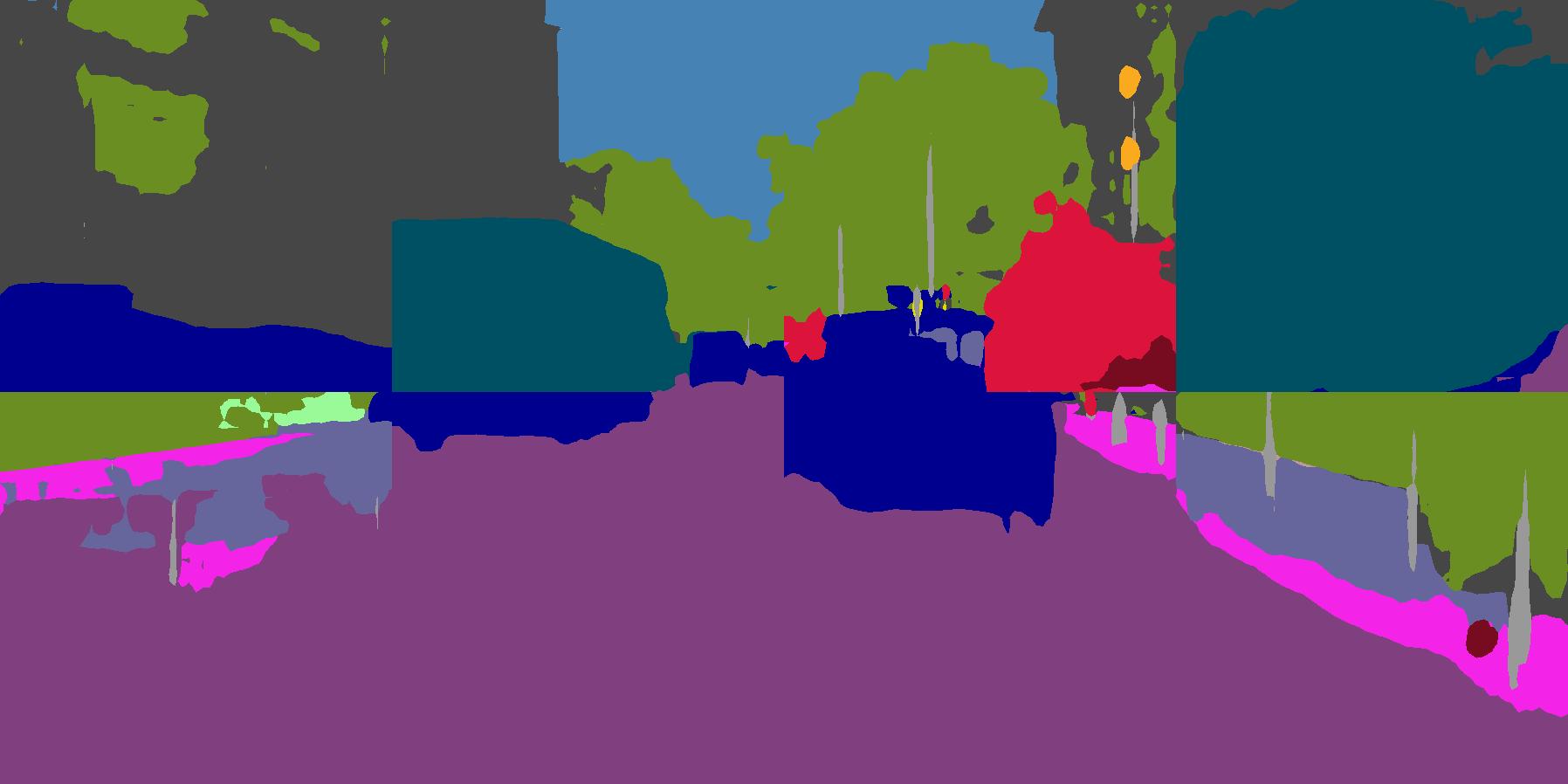}\\
\end{tabular}
\caption{Visual illustration of the attacks investigated in \cref{sec:exp-bounded}.
The sample image is the same as in \cref{fig:examples1}.
Note that the $\ell_\infty$ adversarial perturbations bounded by 0.03 are rather visible yet
the adversarially trained networks are able to produce partially meaningful output.
(Note that the images were processed in non-overlapping blocks, which results in visual artifacts.)}
\label{fig:examples1}
\end{figure}

\textbf{Best models.}
The best models are CIRA+AT-3 and PGD-AT-100. The former has a slightly lower
upper bound on robustness (last column), but a significantly higher clean mIoU value.
\Cref{fig:examples1,fig:examples2} illustrate the visual effects of the attacks, which
helps us put the measurement data in context.

\textbf{Success of CIRA+ on DDC-AT and PGD-AT-50.}
It is clear that a high iteration number results in a much stronger attack
in all the cases.
More importantly, an aggressive attack by CIRA+, using 120 iterations, achieves an almost
perfect attack success \emph{reducing the mIoU score to almost zero on PGD-AT and DDC-AT}.

\textbf{Benefit of uncorrelated attacks.}
It is clear that on different models different attacks are successful, which
strongly supports the idea that one should use a diverse set of attacks for evaluation.
For example, one could expect the attack that was used during adversarial training to be
relatively less successful than other attacks.
Indeed, on the CIRA+AT models the PGD attack is more successful than
the CIRA+ attack.
The same effect can be observed over the PGD-AT and DDC-AT models with CIRA being more
successful than PGD.

\textbf{100\% vs. 50\% adversarial batches.}
Interestingly, PGD-AT-100 is less robust to the very attack it was trained on (ie., PGD)
than PGD-AT-50, even though the latter model has seen fewer adversarial samples.
Yet, PGD-AT-100 is a lot more robust to the CIRA types of attacks than even the CIRA+AT
models.
It looks as though half-batch adversarial training introduces gradient
obfuscation that disrupts the PGD attack specifically, but using a fully adversarial batch
removes this obfuscation. This is an interesting hypothesis to be investigated further
in the future.

\textbf{Robustness accuracy trade-off.}
We can observe the robustness accuracy trade-off in our set of models, a phenomenon
that is well-known in image classification~\cite{Tsipras2019a}.
This also supports the rule of thumb that if a model has a large clean accuracy then
it is less likely to be actually robust.

\section{Minimum Perturbation Attacks}
\label{sec:almaprox}

Here, we evaluate our model instances using a set of attacks that follow a different
goal: they want to minimize the adversarial perturbation for a given level of error.
One example of such an attack is ALMAProx~\cite{Rony2022a}.

\subsection{Definition of the Minimum Perturbation Attack}

Instead of
maximizing the distortion of the output mask within a bounded set of input
perturbations $\Delta$,
another possible attack model is to set a fixed amount of distortion and
minimize the size of the perturbation to achieve it.

To formulate such a problem, we define the fixed amount of distortion to be
the pixel error, that is, a fixed proportion of pixels that are misclassified by a
model $f_\theta$ given the training sample $(x,y)$.
We will denote this quantity by PE$(f_\theta(x),y)$.
The corresponding optimization problem can be formulated as
\begin{equation}
\delta^*=\arg\min_{x+\delta\in\mathcal X} \|\delta\|_\infty\ \ \mathrm{s.t.\ PE}(f_\theta(x+\delta),y)>=\mu,
\label{eq:minperturb}
\end{equation}
where $\mu$ is a constant defining the desired pixel error level, and
$(x,y)$ is the labeled input example being attacked.
Obviously, this constrained minimization problem is not guaranteed to
have a feasible solution, but---since all the possible inputs are allowed---a feasible
solution is expected to exist.
We also note that Rony et al.~\cite{Rony2022a} ignore those pixels that are labeled
with the ``void'' class, and measure pixel error only over the rest of the pixels.
We adopt this practice here.

\begin{figure*}[tb]
\centering
\includegraphics[width=0.32\textwidth]{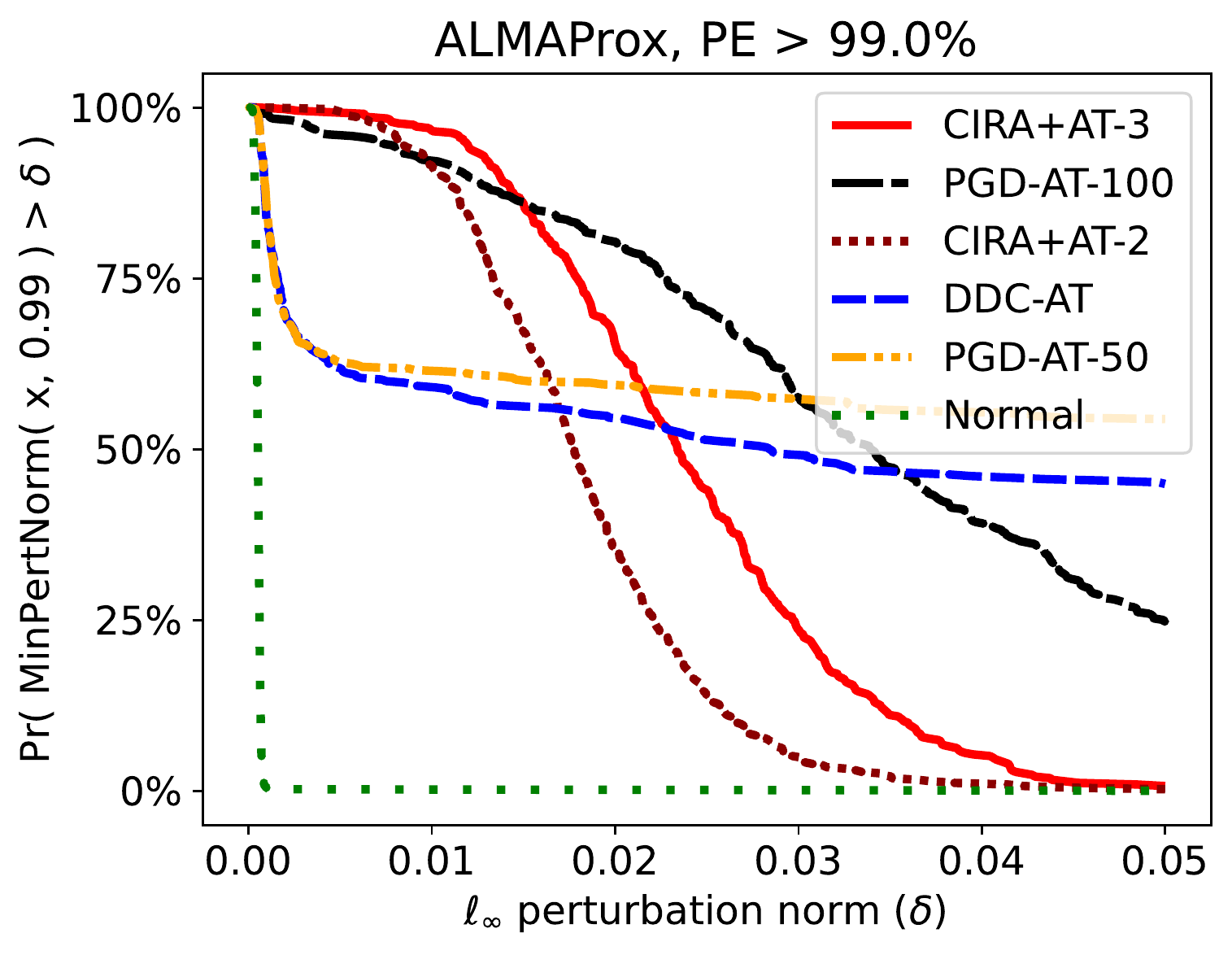}
\includegraphics[width=0.32\textwidth]{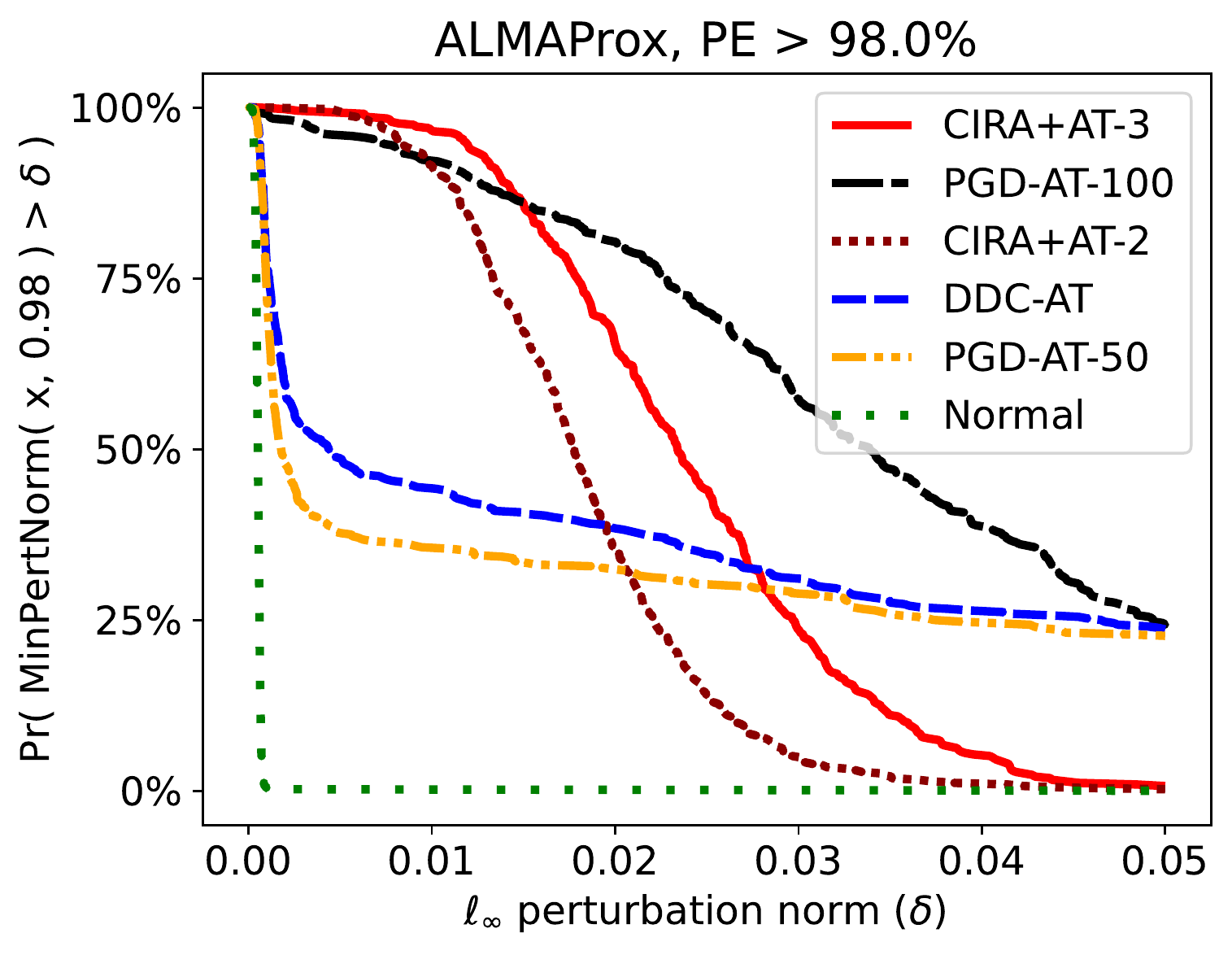}
\includegraphics[width=0.32\textwidth]{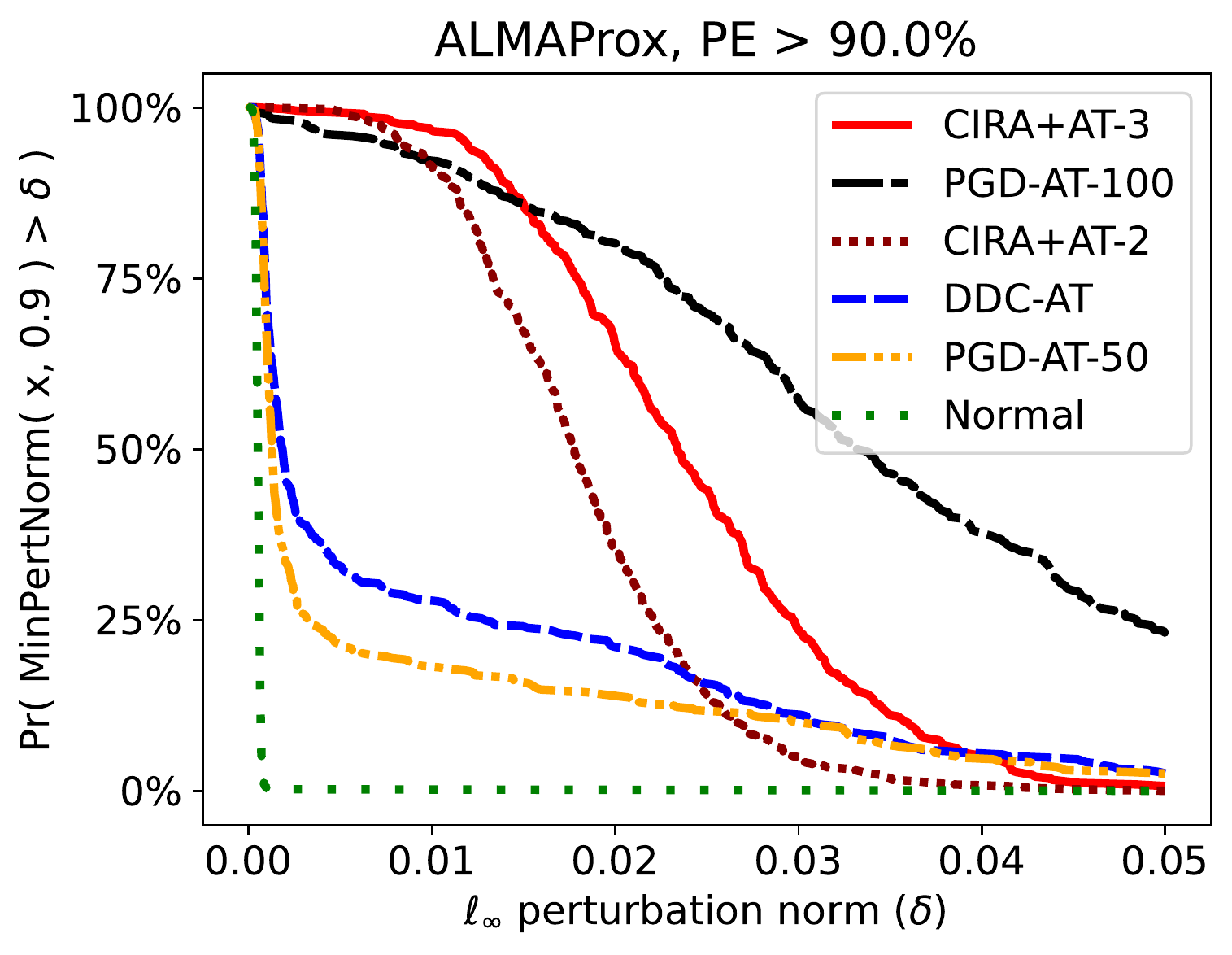}
\includegraphics[width=0.32\textwidth]{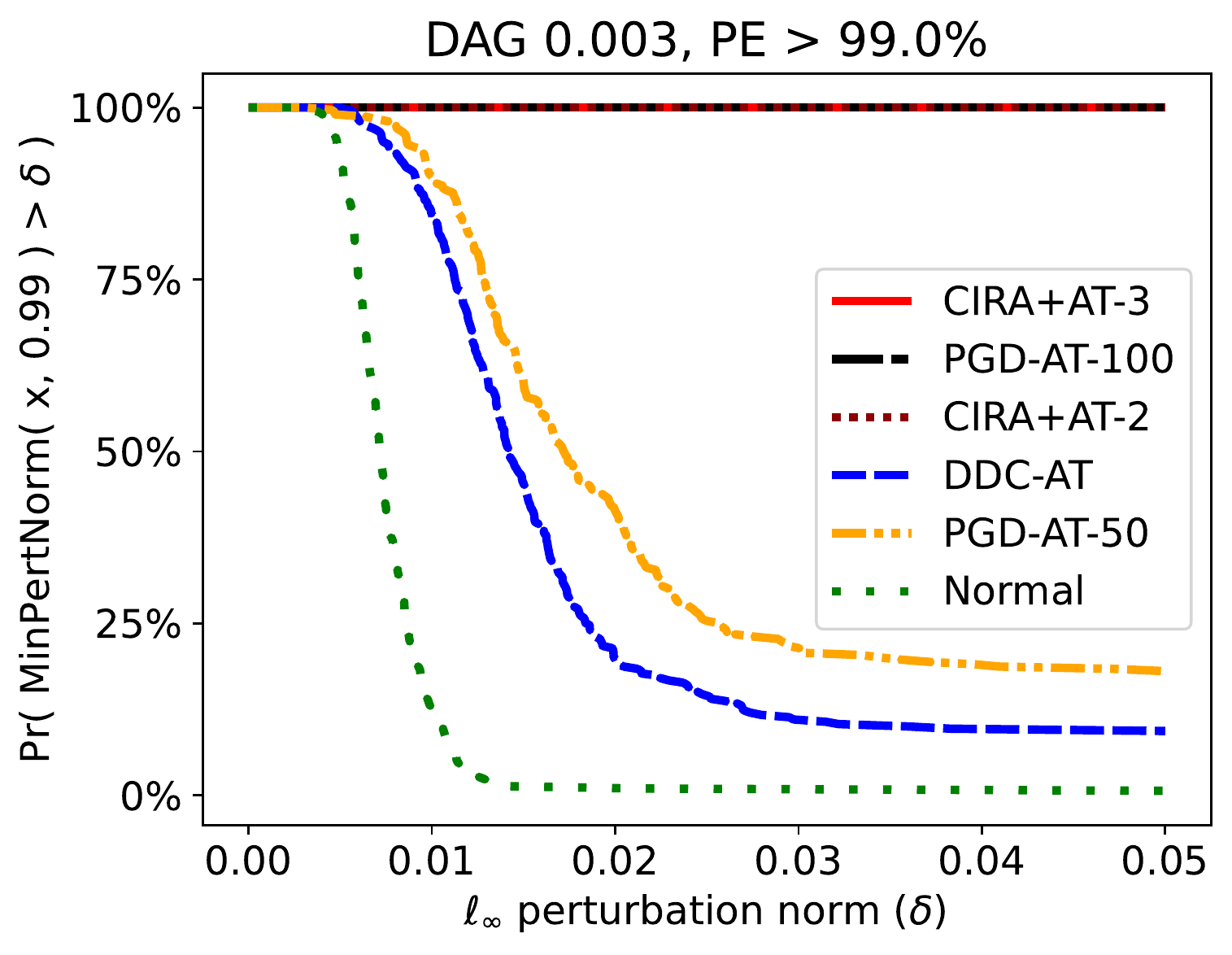}
\includegraphics[width=0.32\textwidth]{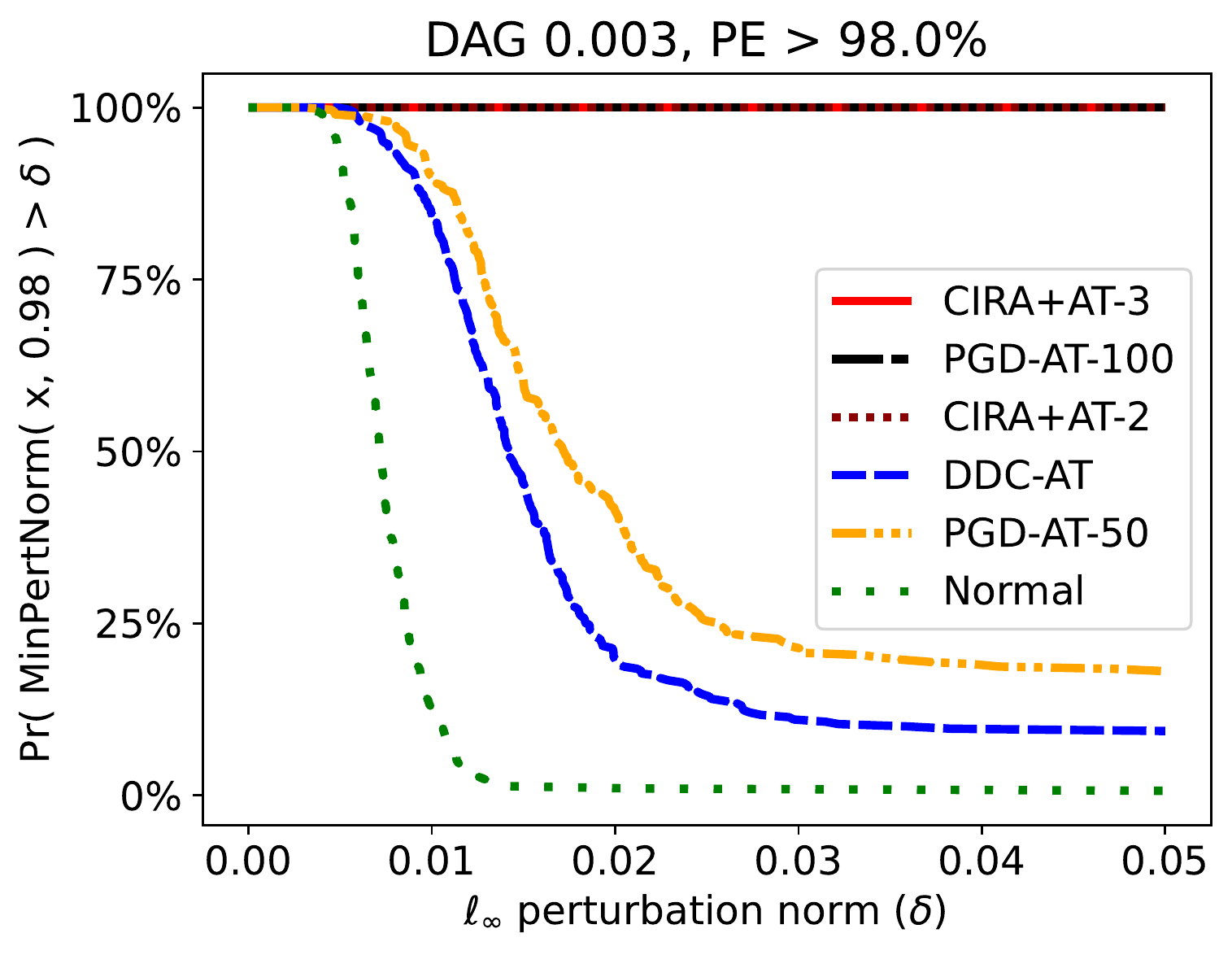}
\includegraphics[width=0.32\textwidth]{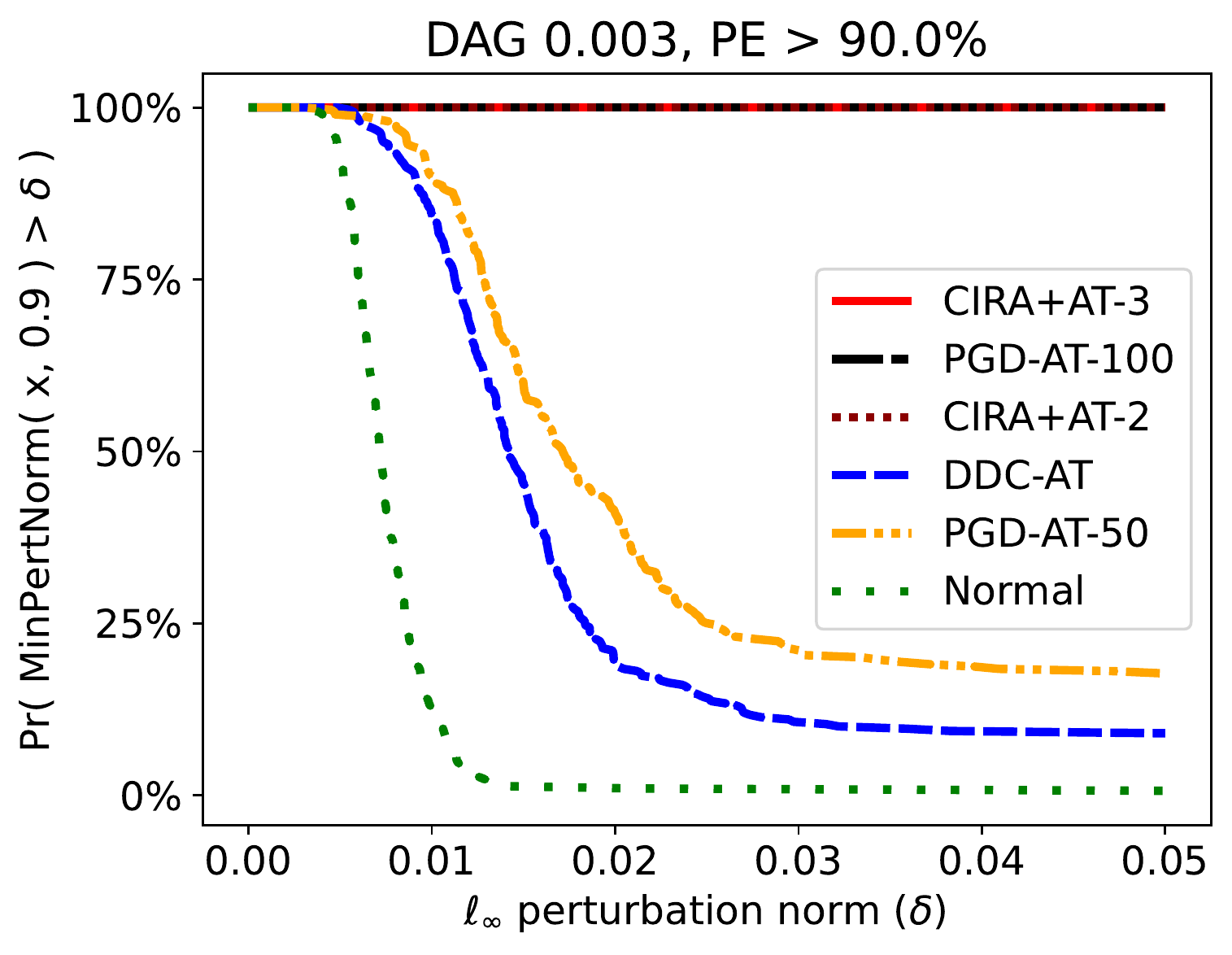}
\includegraphics[width=0.32\textwidth]{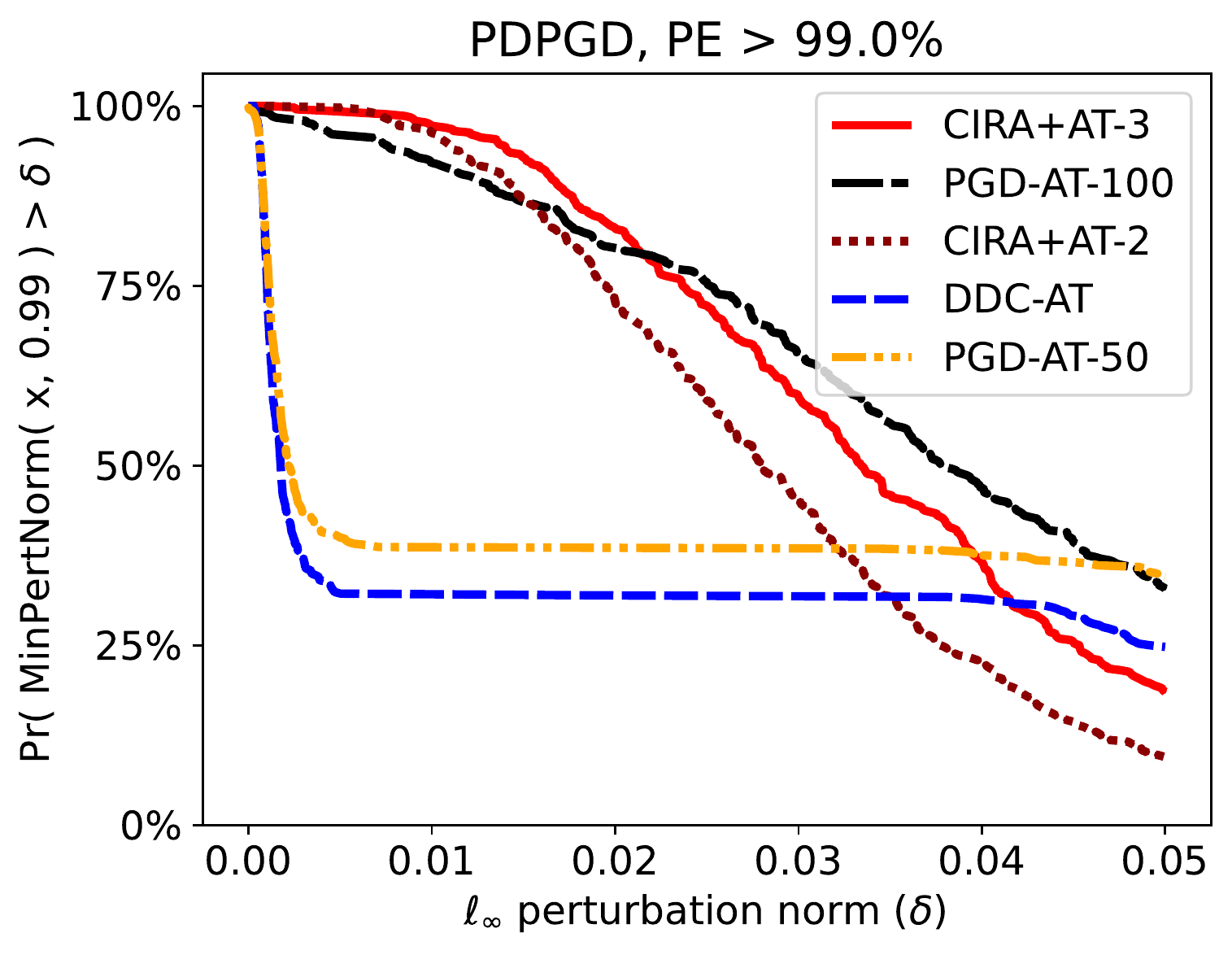}
\includegraphics[width=0.32\textwidth]{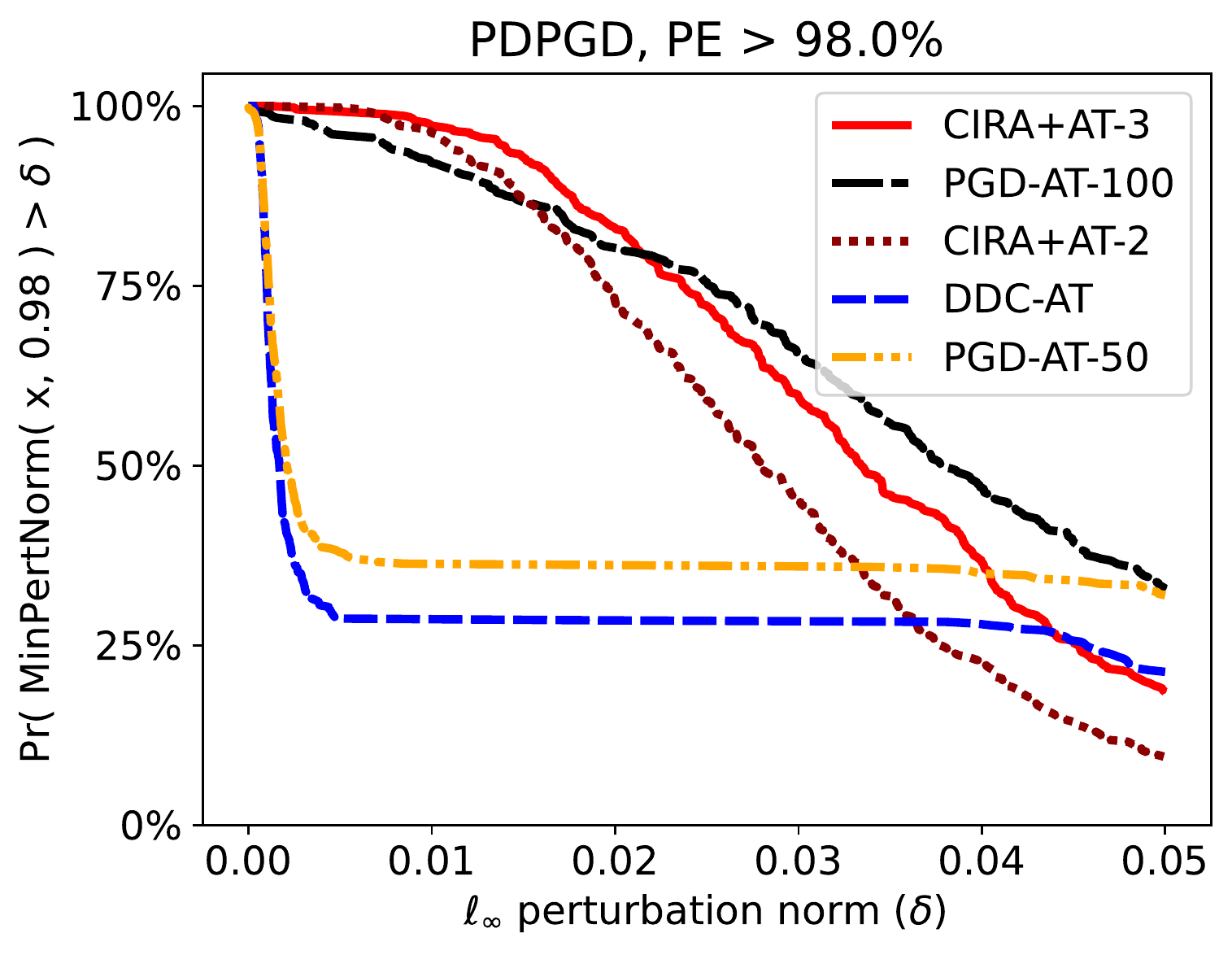}
\includegraphics[width=0.32\textwidth]{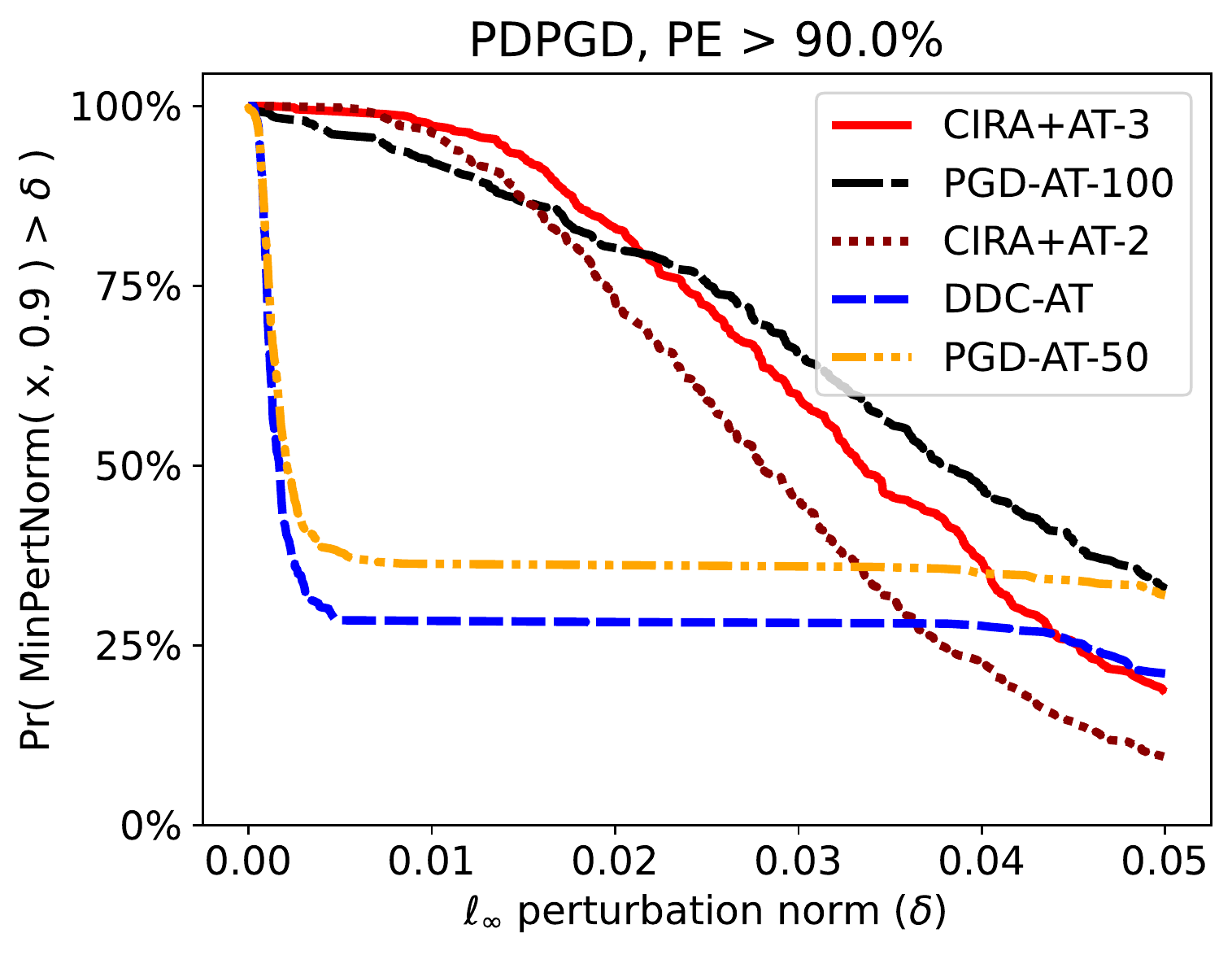}
\caption{The plots show the empirical probability that the $\ell_\infty$ norm of the
minimal perturbation to achieve a given pixel error on a random input $x$ is larger than a
given value.
Three attacks are shown: ALMAProx (top row), DAG with step-size 0.003 (middle row) and
PDPGD (bottom row).}
\label{fig:pspnet-apsr}
\end{figure*}

\begin{figure*}[tb]
\centering
\includegraphics[width=0.32\textwidth]{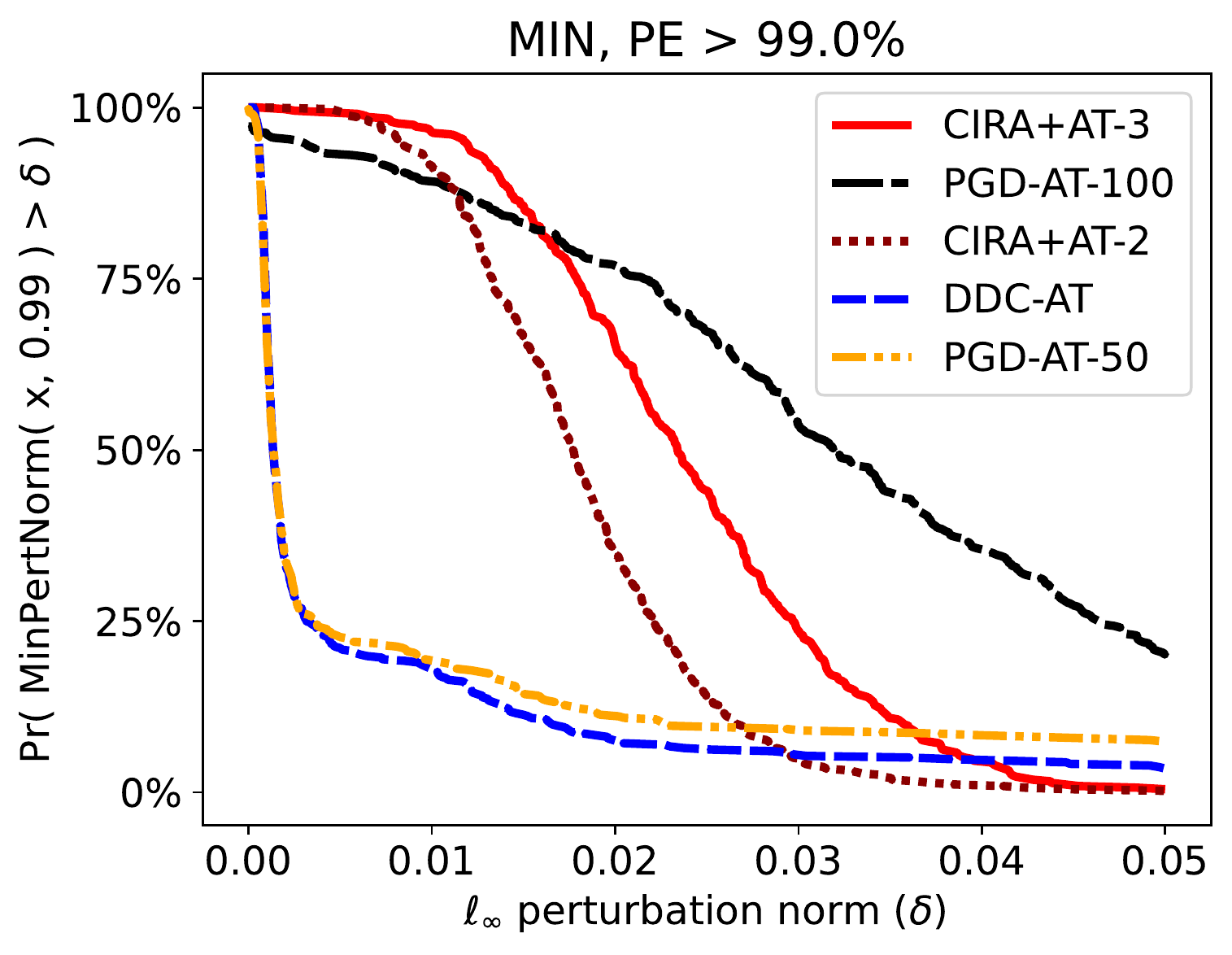}
\includegraphics[width=0.32\textwidth]{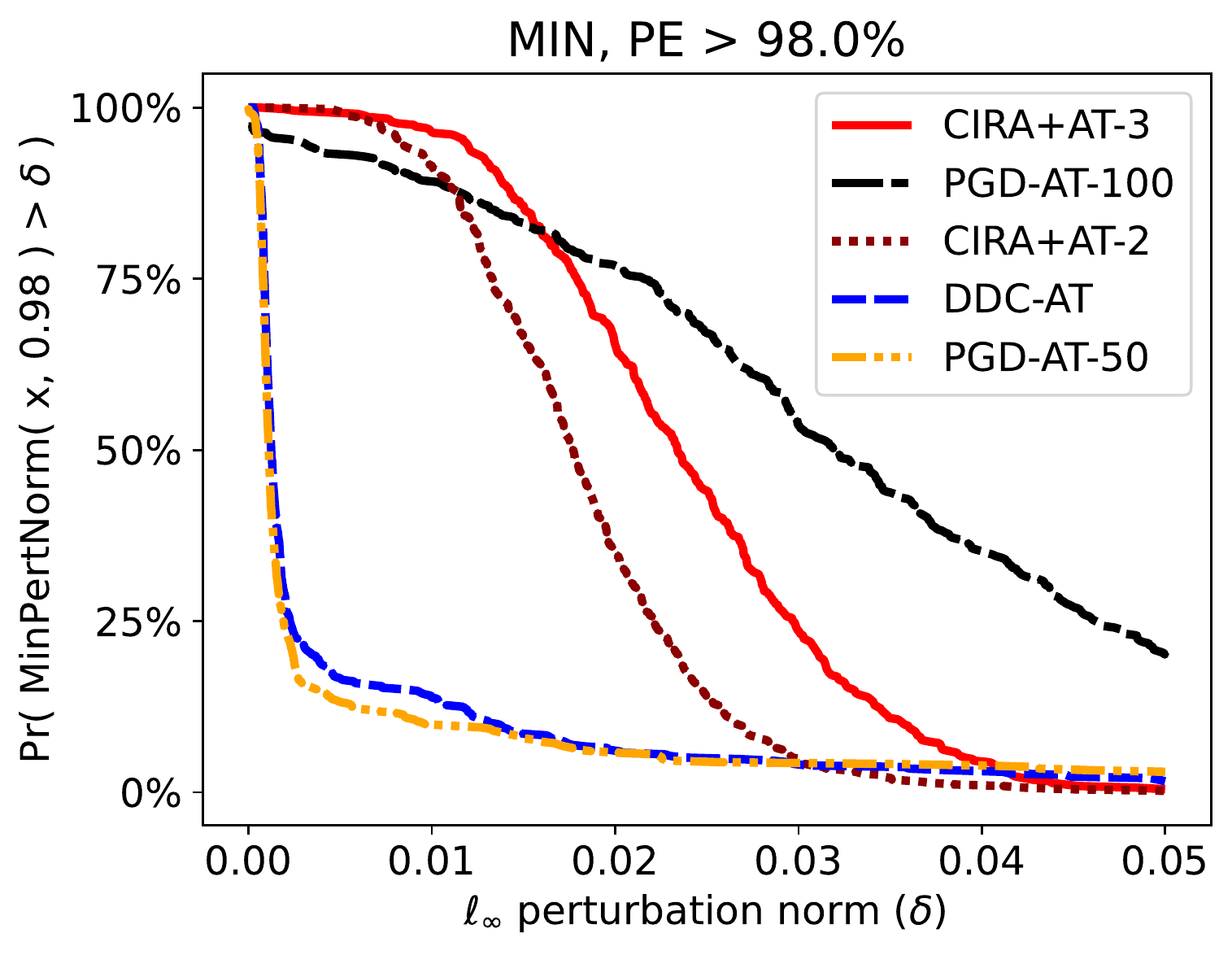}
\includegraphics[width=0.32\textwidth]{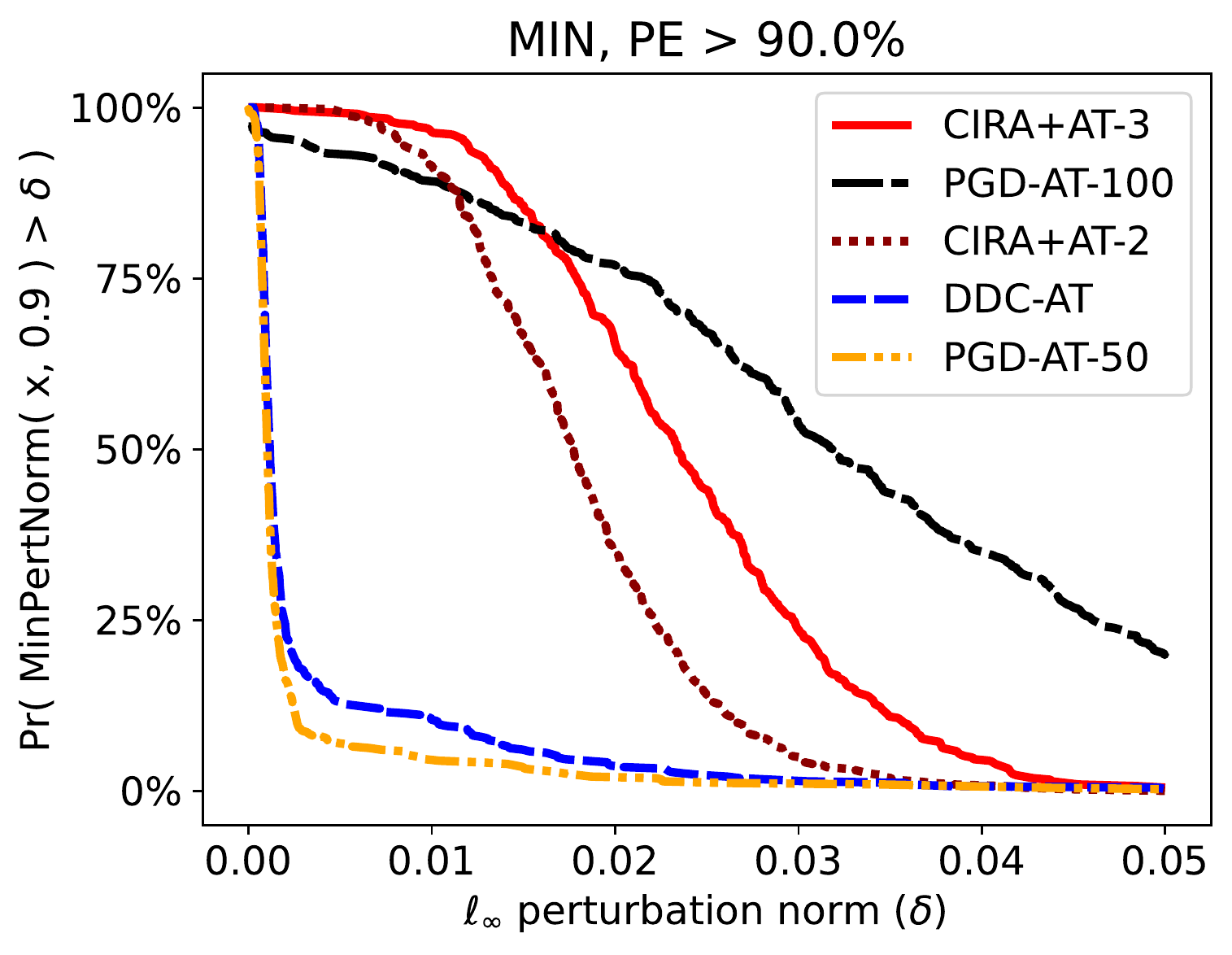}
\caption{The plots show the empirical probability that the $\ell_\infty$ norm of the
minimal perturbation to achieve a given pixel error on a random input $x$ is larger than a
given value.
The plots illustrate the aggregated minimum perturbations over all the attacks.}
\label{fig:pspnet-merge}
\end{figure*}

\subsection{Evaluation Methodology}

When evaluating the robustness of a model,
we execute a number of different attacks on the model to approximate the minimal perturbation
required to achieve a pixel error of 99\% over a test set.
For each test example and for each attack,
we record the minimal perturbation found along with the corresponding pixel error.
Using this data, we can present the distribution of the norm of the minimal perturbation
for the attacks.

Note, that---although we set a 99\% target---the attacks we include here
can report lower percentages if they fail to reach 99\%.
Thus, this data allows us to present the distribution of minimal perturbations for
not only 99\%, but also any lower level $\mu$ of pixel error.
To do this, we treat all the
perturbations that could not reach a pixel error of $\mu$ to have an infinite norm.

\subsection{Attacks}

\textbf{ALMAProx.} 
Proposed by Rony et al.~\cite{Rony2022a}, the ALMAProx attack is a 
proximal gradient method for solving the perturbation
minimization problem in \cref{eq:minperturb}.
The problem is transformed into an unconstrained problem by moving the constraints
into the objective using Lagrangian penalty functions.
The parameter settings we used were identical to those in~\cite{Rony2022a}.

\textbf{DAG.} Xie et al.\ proposed the dense adversary generation (DAG)
algorithm~\cite{Xie2017a}, which attempts to attack each pixel using a
gradient method that takes into consideration only the correctly predicted
pixels in each gradient step.
The method is not bounded (the perturbation can grow without bounds),
nor does it minimize the perturbation explicitly.
Instead, the algorithm terminates when reaching the maximum iteration number, or
when all the targeted pixels are successfully attacked.
We can use this algorithm in our evaluation framework by recording the
perturbation size at termination, along with the ratio of the successfully attacked
pixels.
The maximum iteration number was set to 200, and we used two step-sizes:
0.001 and 0.003.

\textbf{PDPGD.}
Matyasko and Chau proposed the primal-dual proximal gradient descent adversarial attack
(PDPGD)~\cite{Matyasko2021a} that,
like ALMAProx, also uses proximal splitting but uses a different
optimization method.
PDPGD was adapted to the semantic segmentation task in~\cite{Rony2022a} via
introducing a constraint on each pixel and ignoring the pixels with the ``void'' label.
We applied the same parameter settings as~\cite{Rony2022a}.

\subsection{Results}

\Cref{fig:pspnet-apsr} shows the distribution of the minimum perturbations
for three attacks (ALMAProx, DAG, PDPGD) individually, and \cref{fig:pspnet-merge} shows
the aggregated minimal perturbation, where for each image we select the minimum perturbation
found by any of the successful attacks.
Note that $\ell_\infty$ perturbations with a norm smaller than about $0.015$ are
practically invisible to the human eye.

The normal model has no robustness at all as shown by the ALMAProx results so,
for clarity, we show the results on the normal model only for ALMAProx and DAG.

\textbf{Best models.}
The best models are again CIRA+AT-3 and PGD-AT-100.
Here, we can see that PGD-AT-100 is more robust, in that a very high percentage
of examples have a much higher upper bound on the minimal perturbation that
is required to achieve 99\% pixel error. 
However, there are somewhat fewer images that require only a very small perturbation,
when using CIRA+AT-3.
Recall, that beyond a perturbation norm of $0.015$ the perturbation is becoming visible.
Thus, in the ``invisiblity range'' of $[0,0.015]$ CIRA+AT-3 is superior.

\textbf{Benefit of uncorrelated attacks.}
Here, we can see further evidence that it is strongly advisable to use uncorrelated
attacks.
While the DAG attack is rather weak, it still provides a very valuable contribution
to the aggregate evaluation, because in the case of DDC-AT and PGD-AT-50 it is
the most successful attack, besides, it can attack different inputs than
ALMPAProx or PDPGD.
This provides further evidence
that DDC-AT and PGD-AT-50 are not robust, but the CIRA+ models and PGD-AT-100 might be.

\textbf{100\% vs. 50\% adversarial batches.}
Like in the case of the bounded attacks, here we can also observe a peculiar
feature of the models that use 50\% adversarial batches.
Namely, the ALMAProx algorithm is not able to achieve 99\% pixel error at all
on a significant proportion of test images,
while it can do so (although with large perturbations)
in the case of all the 100\% adversarially trained models.
This, again, seems to contradict the fact that the 50\% models are less robust
overall, when taking into account the entire set of attacks.
This is consistent with our speculative hypothesis that these models have obfuscated
gradients that, in this case, disturb ALMAProx more than DAG.

\section{Limitations and Conclusions}

Due to putting a large emphasis on the thoroughness of the robustness evaluation and
the limited amount of computational resources we have access to, we could
investigate only a handful of model instances, which at this time
prevents us from drawing fine-grained
conclusions about the effect of the different hyperparameters and design decisions
on the robustness of the models.
The model instances we discussed were intended primarily as a constructive proof
for our claimed contributions regarding the evaluation methodology.

We have demonstrated that it is very important to utilize a diverse set of hard
attacks to evaluate robustness.
Also, we have seen that the aggregation of the results of several attacks
can sometimes provide a much lower upper bound on the robustness than any
attack individually.

Using this methodology, and with the help of the CIRA+ attack we proposed,
we were able to demonstrate the lack of robustness of some known models, we
proposed simple methods to train models that appear to be robust, and
we demonstrated the robustness accuracy trade-off over our small set of models.

\section*{Acknowledgements}

This work was supported by the European Union project RRF-2.3.1-21-2022-00004
within the framework of the Artificial Intelligence National Laboratory and
project TKP2021-NVA-09, implemented with the support provided by the Ministry of Innovation and Technology of Hungary
from the National Research, Development and Innovation Fund, financed under the TKP2021-NVA funding scheme and the ÚNKP-22-2-SZTE-343
New National Excellence Program of the Ministry for Culture and Innovation from the source of the National Research, Development and Innovation Fund.

{\small
\bibliographystyle{iccv23/ieee_fullname}
\bibliography{raw,other}
}

\onecolumn
\appendix

\section{Distribution of Best Attacks}

In \cref{sec:almaprox} we argued that it is very important to use a diverse set of attacks, because
different models might be sensitive to different attacks.
This implies that the aggregated minimum robustness is expected to be a
significantly better upper bound on robustness for an unknown model
than that of any of the attacks separately.

\begin{figure}[h]
\centering
\includegraphics[width=0.195\columnwidth]{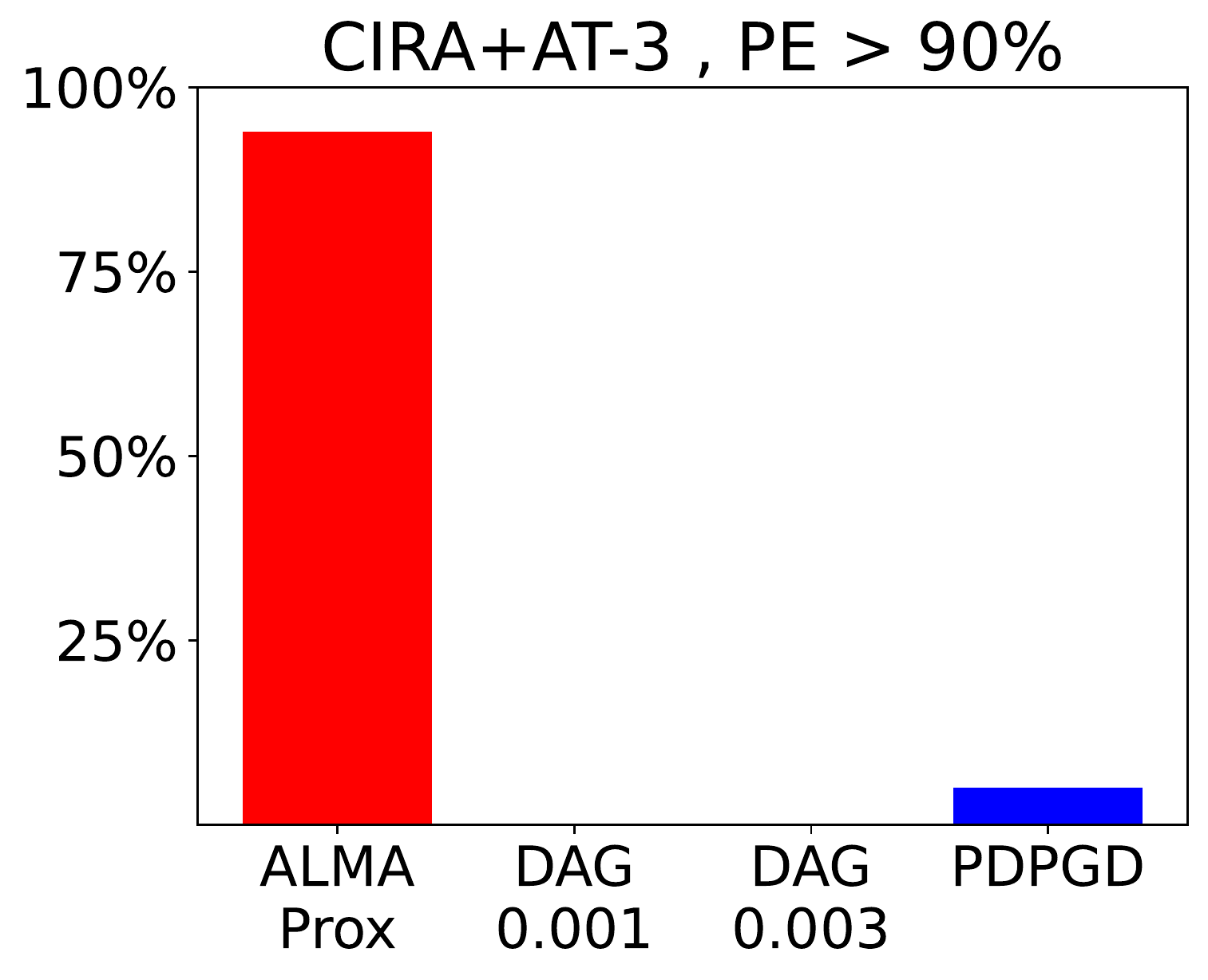}
\includegraphics[width=0.195\columnwidth]{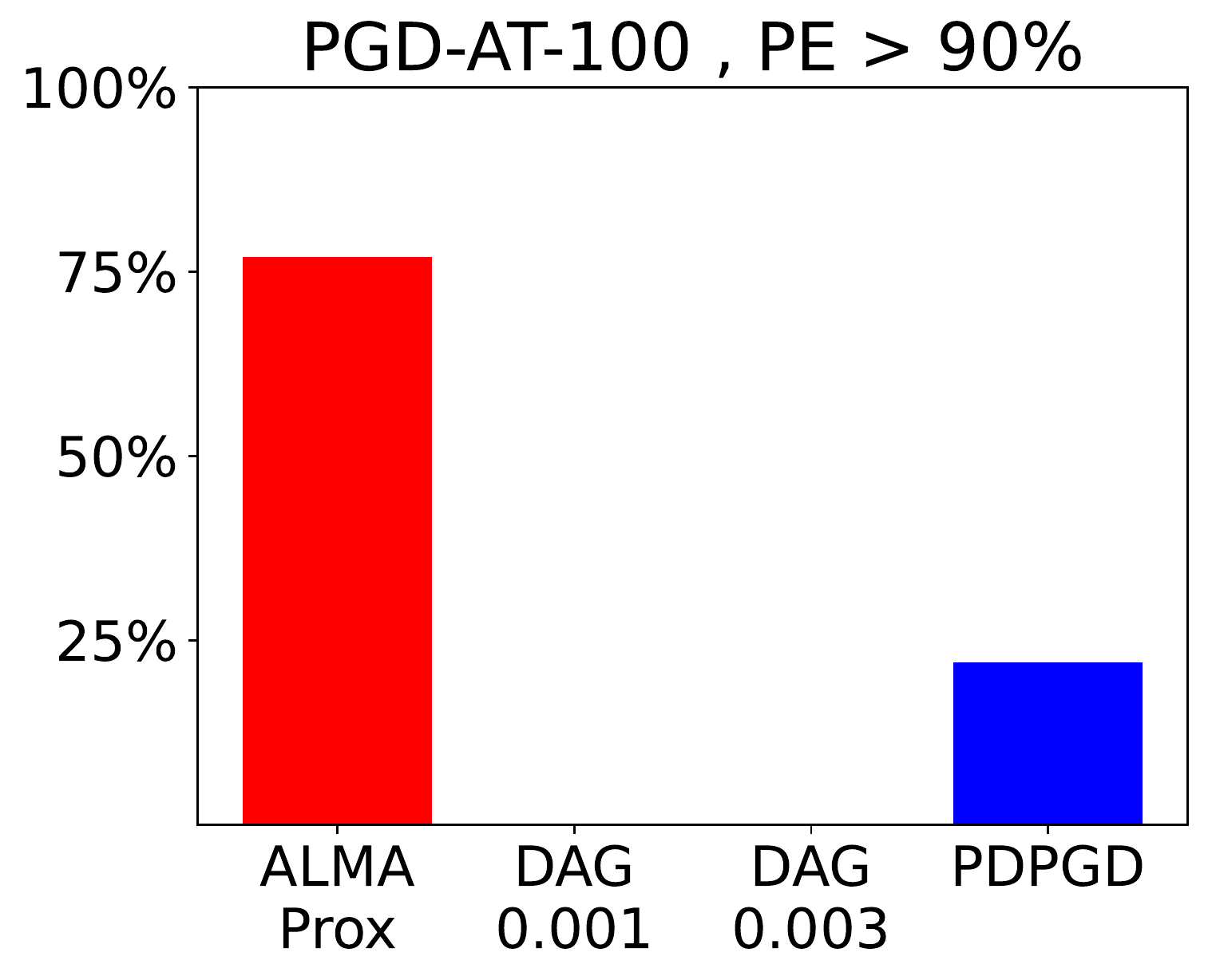}
\includegraphics[width=0.195\columnwidth]{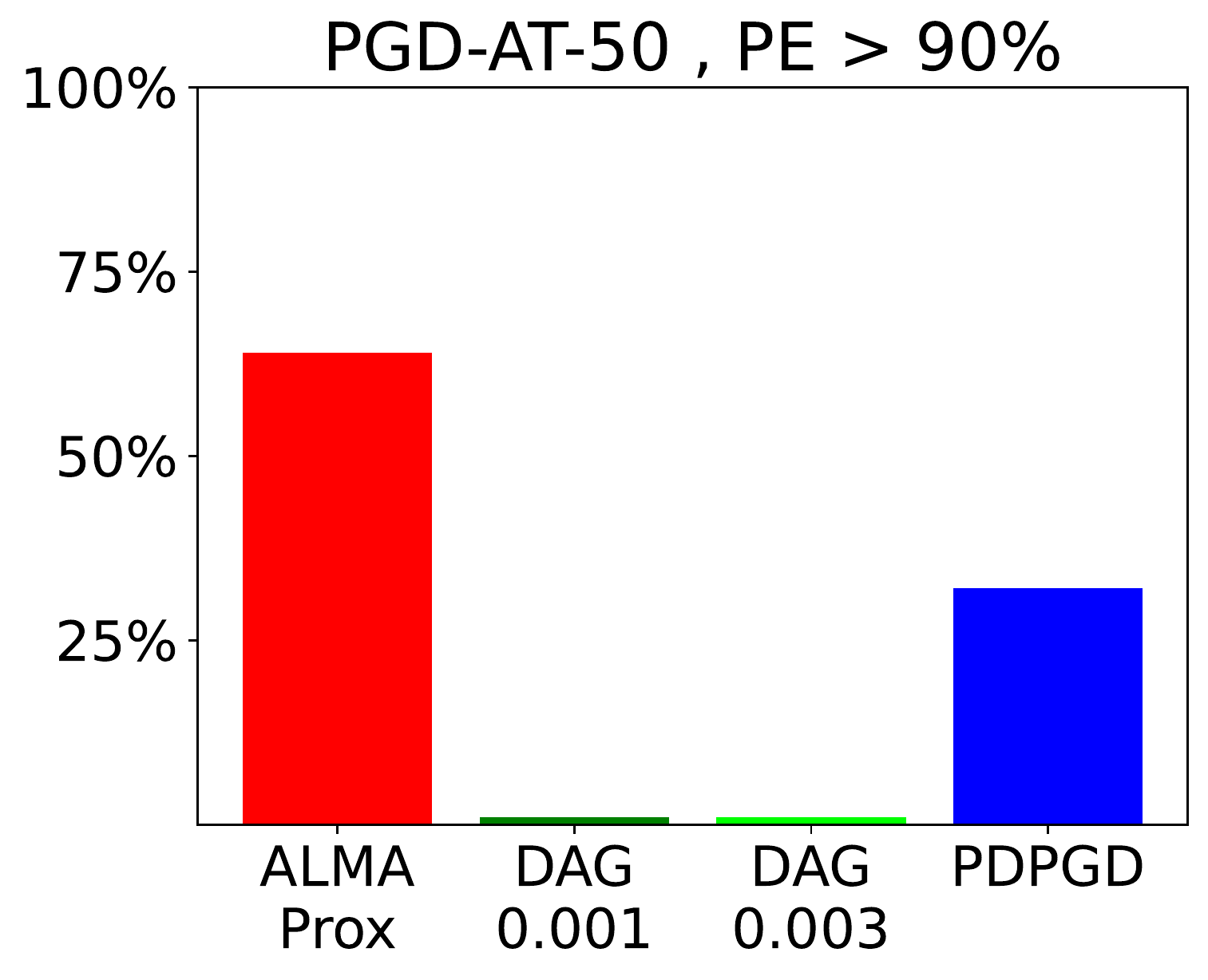}
\includegraphics[width=0.195\columnwidth]{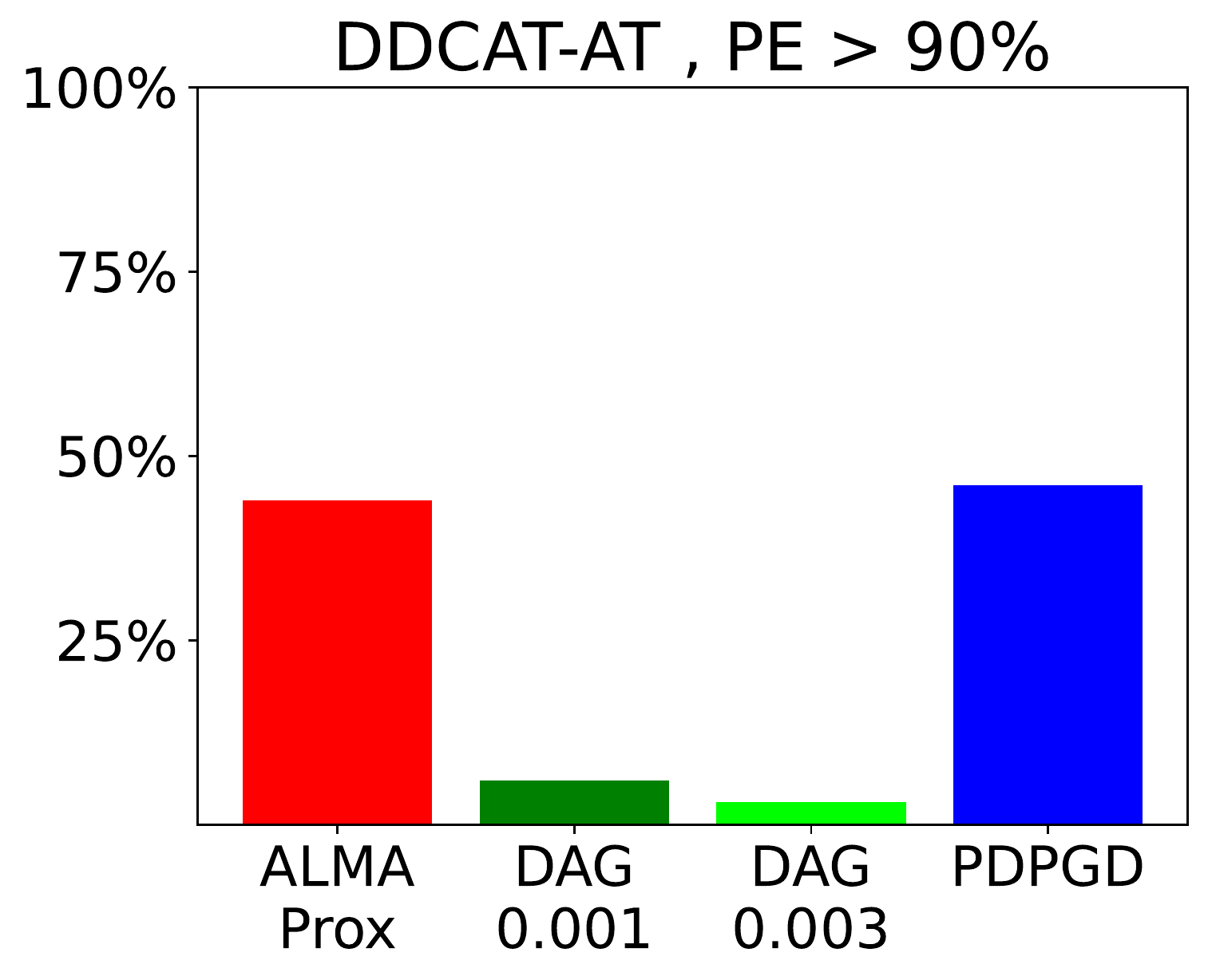}
\includegraphics[width=0.195\columnwidth]{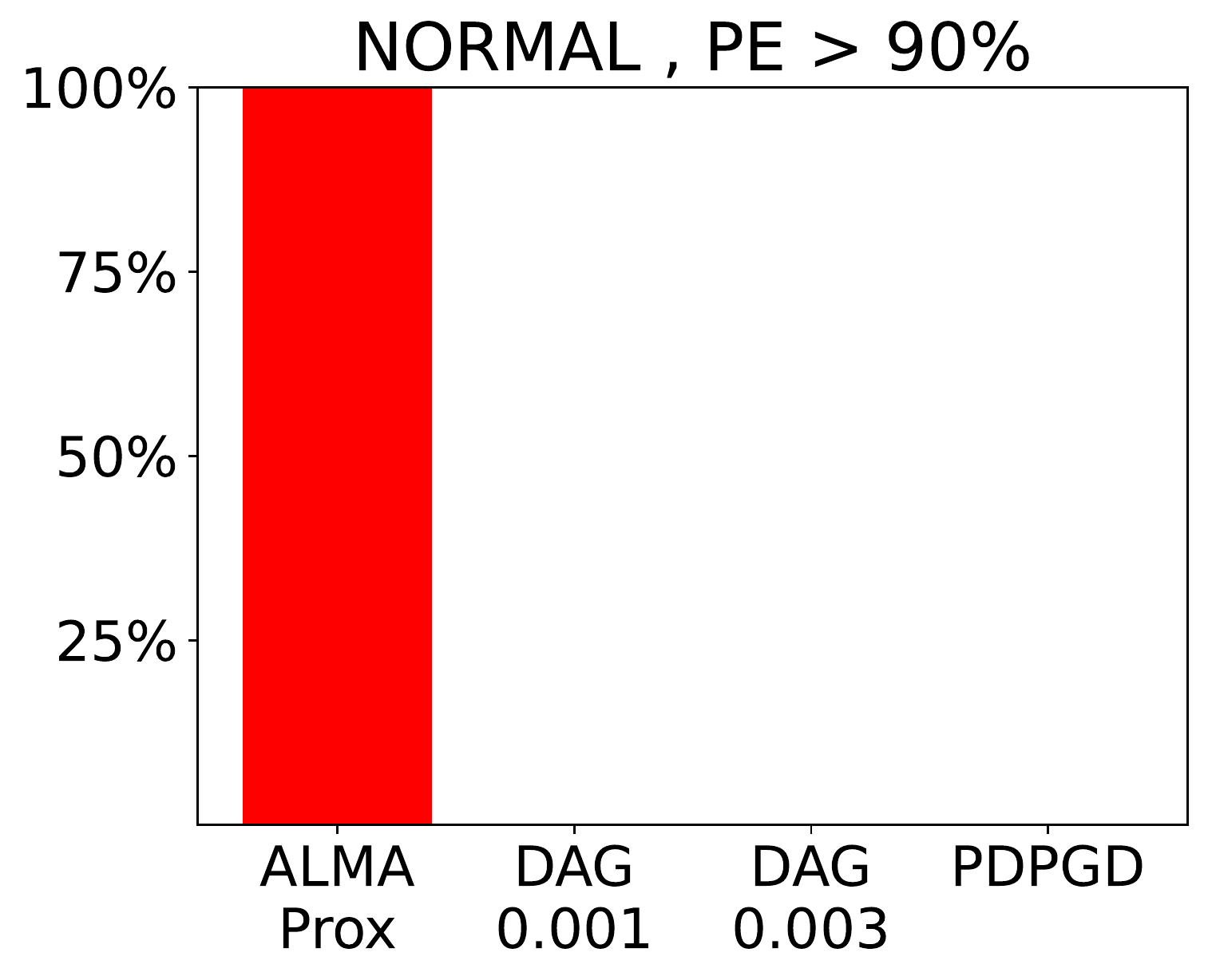}
\includegraphics[width=0.195\columnwidth]{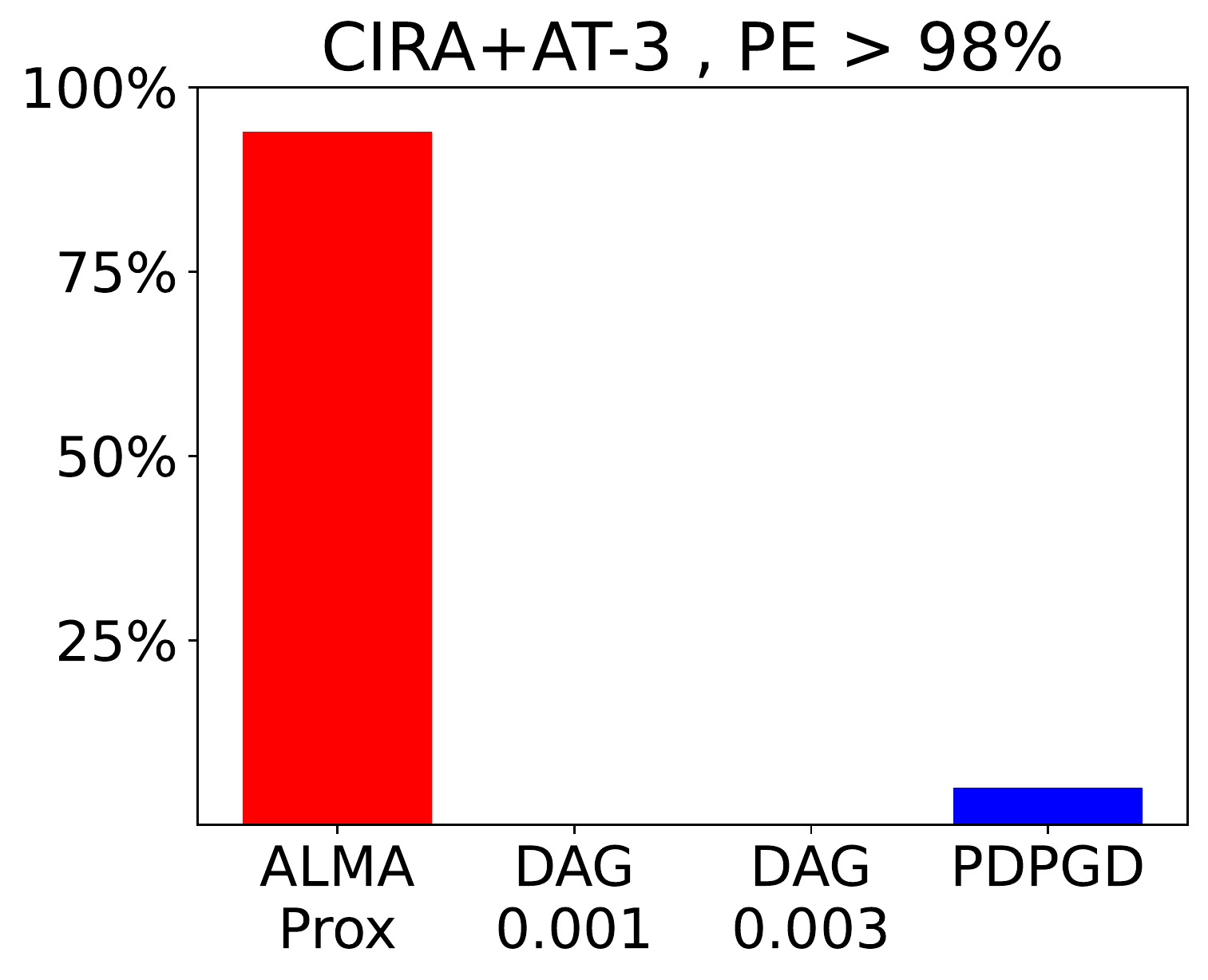}
\includegraphics[width=0.195\columnwidth]{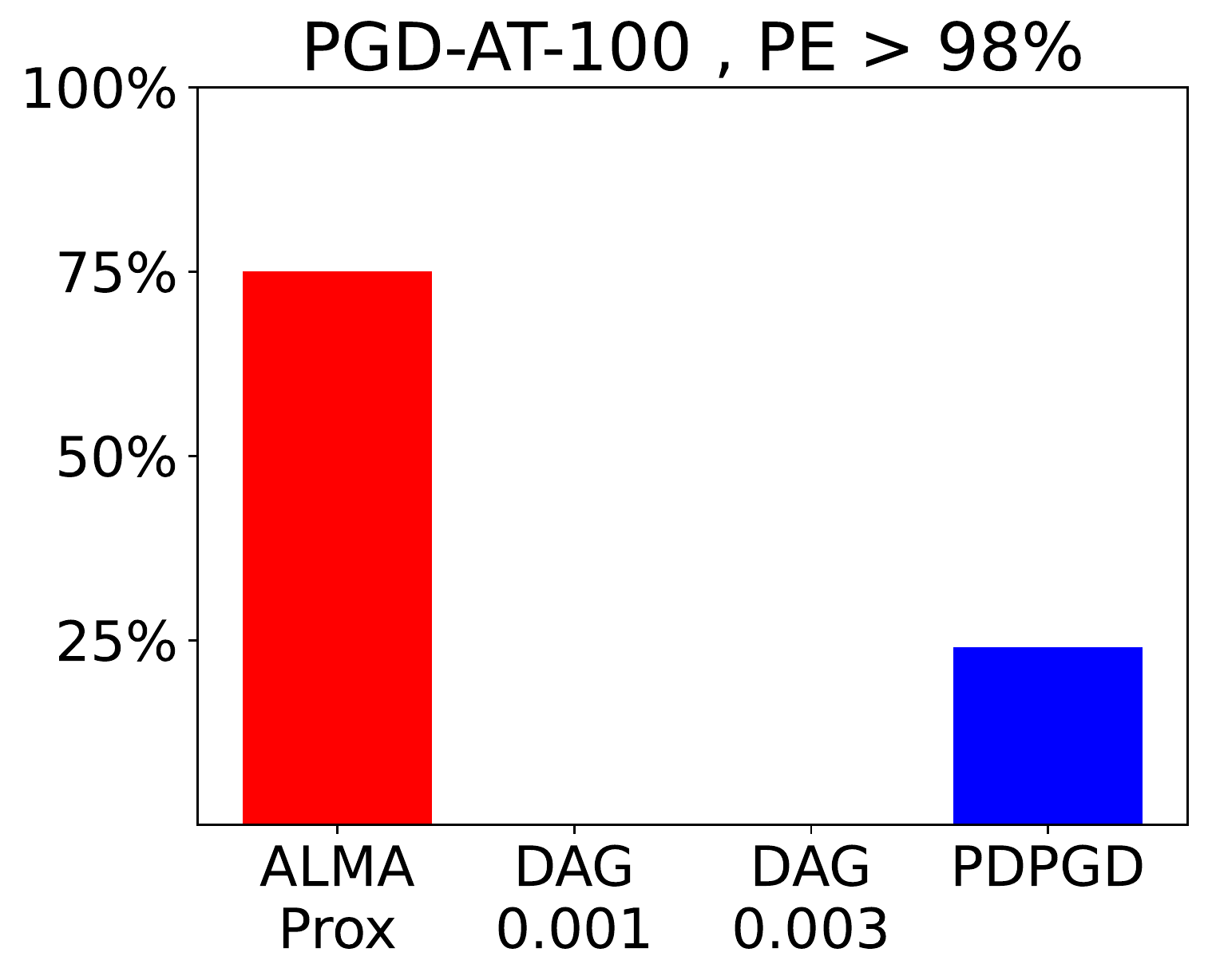}
\includegraphics[width=0.195\columnwidth]{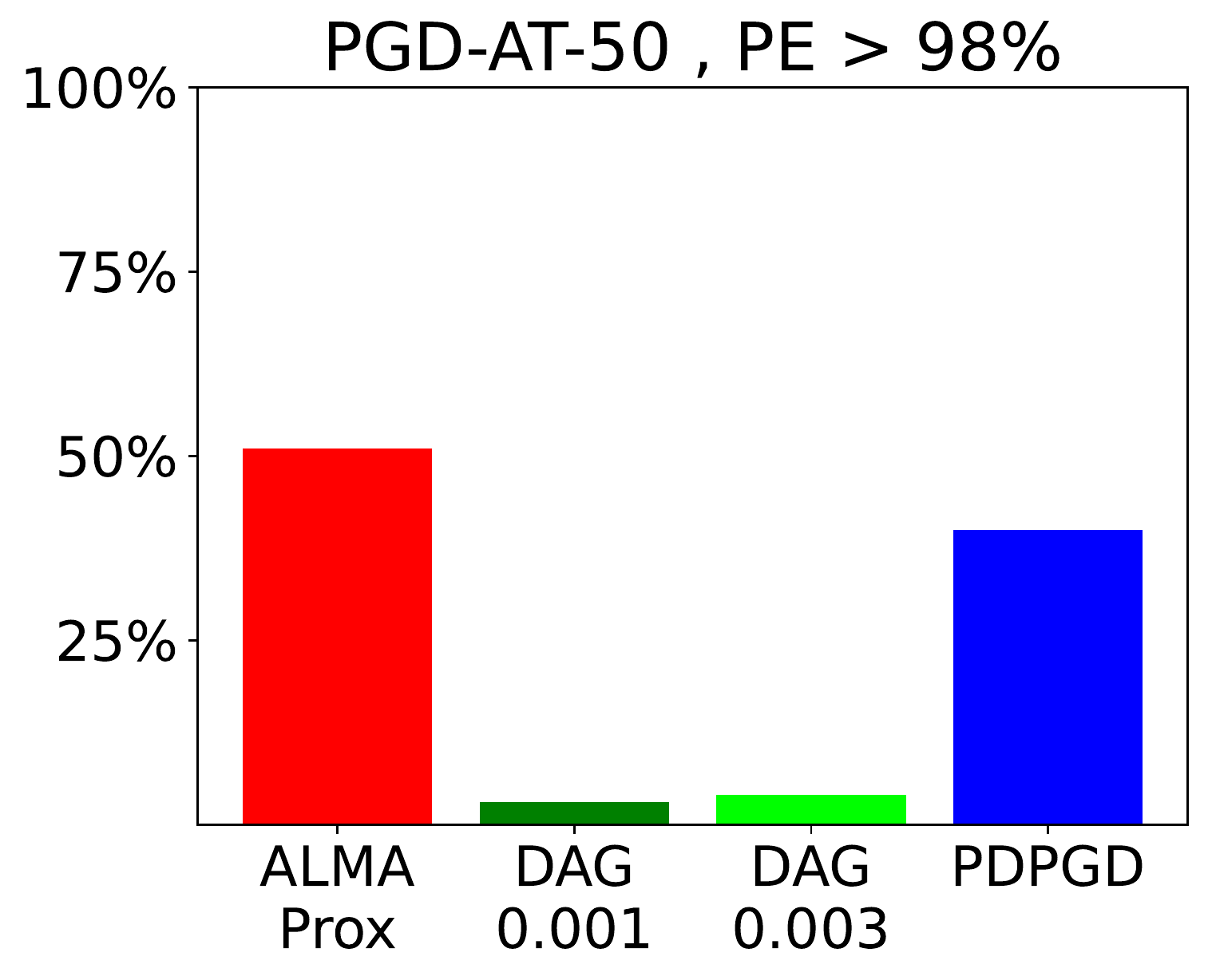}
\includegraphics[width=0.195\columnwidth]{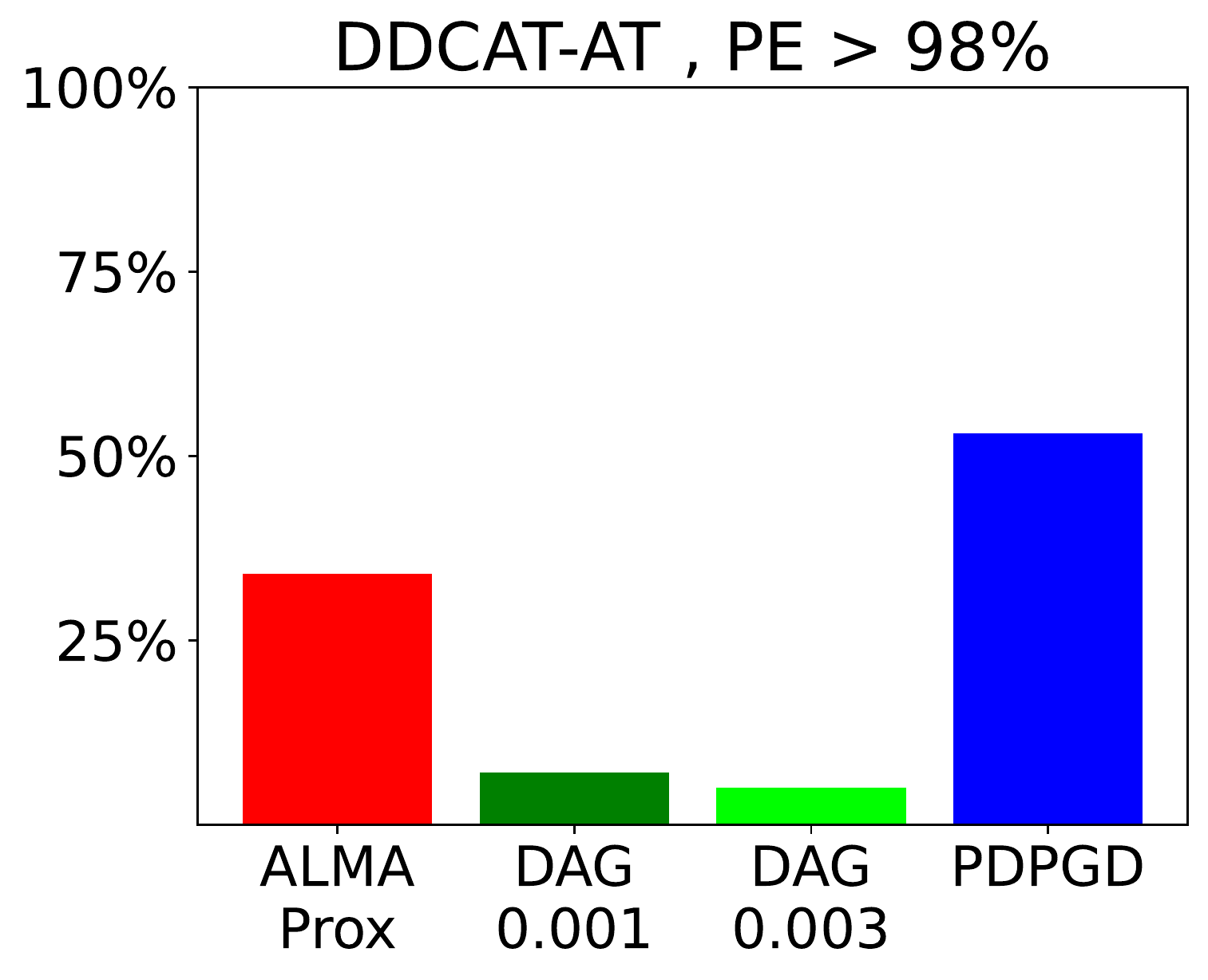}
\includegraphics[width=0.195\columnwidth]{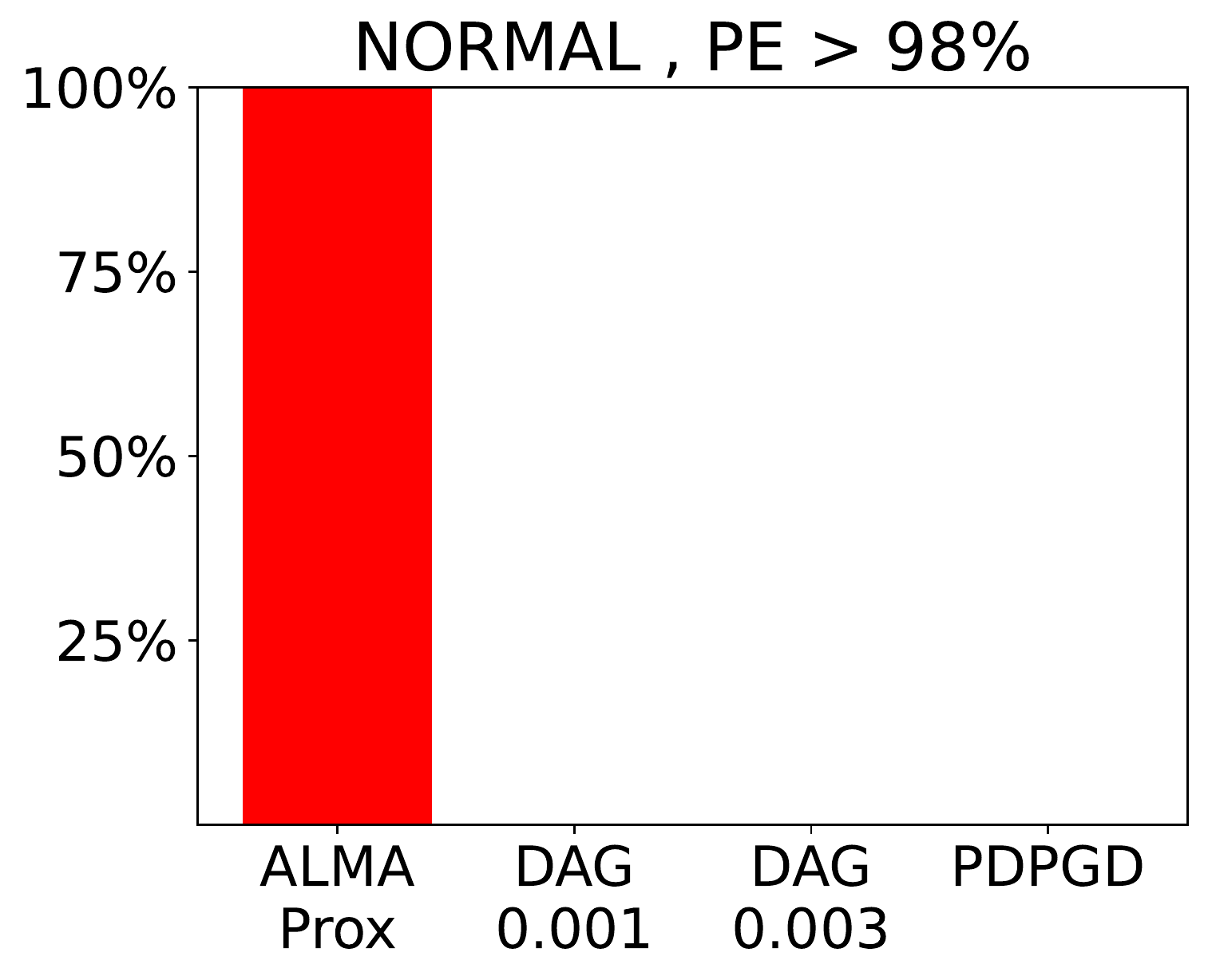}
\includegraphics[width=0.195\columnwidth]{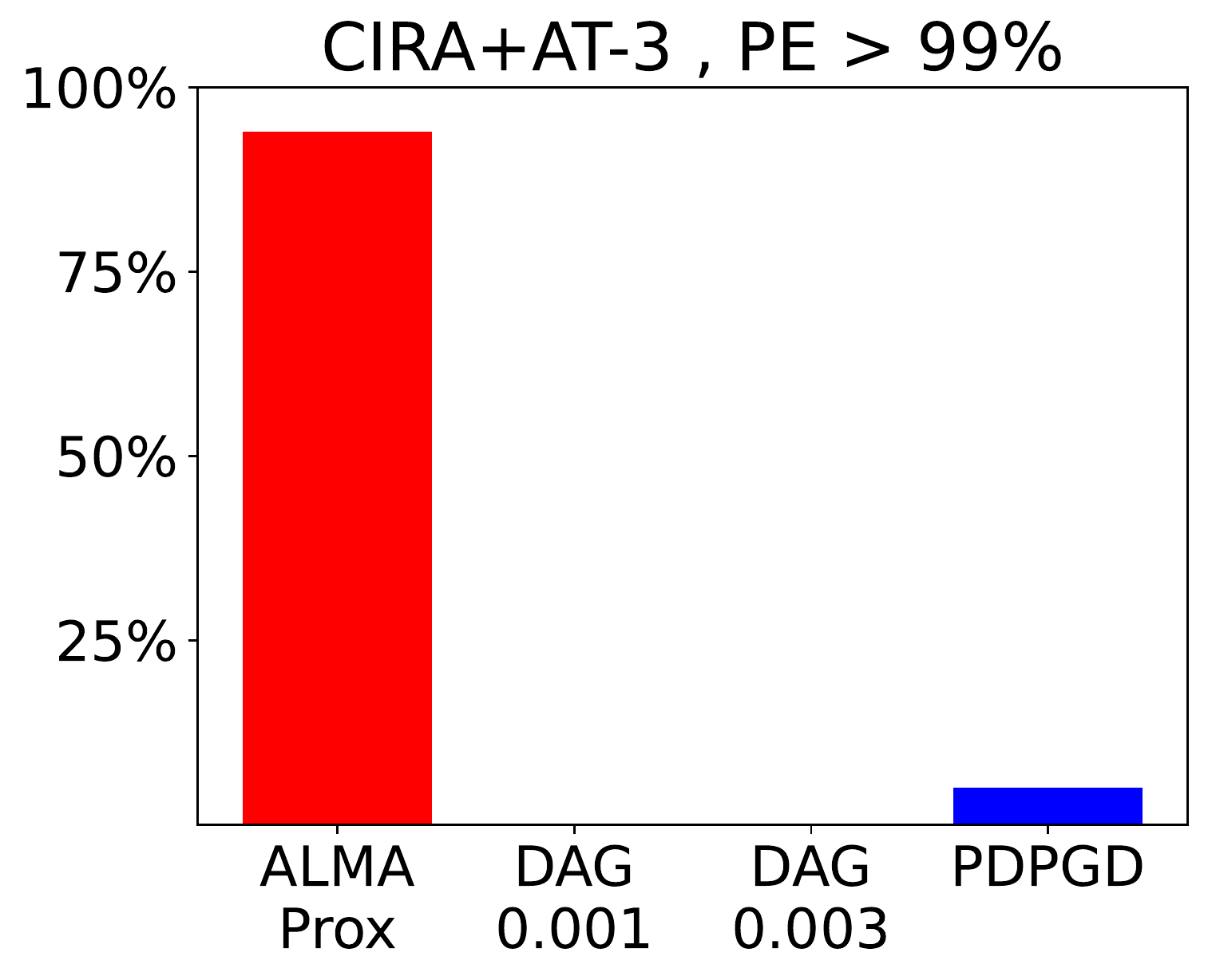}
\includegraphics[width=0.195\columnwidth]{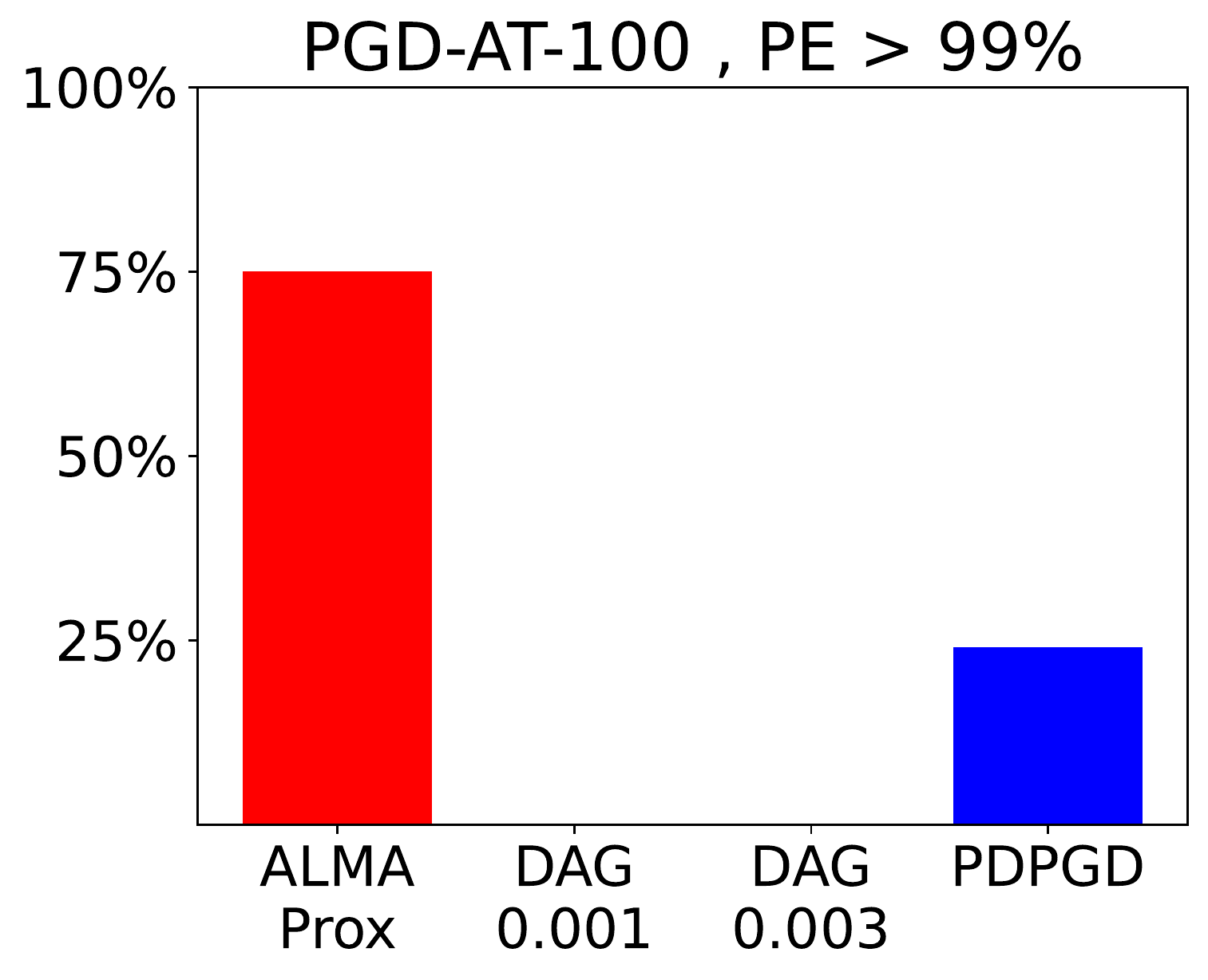}
\includegraphics[width=0.195\columnwidth]{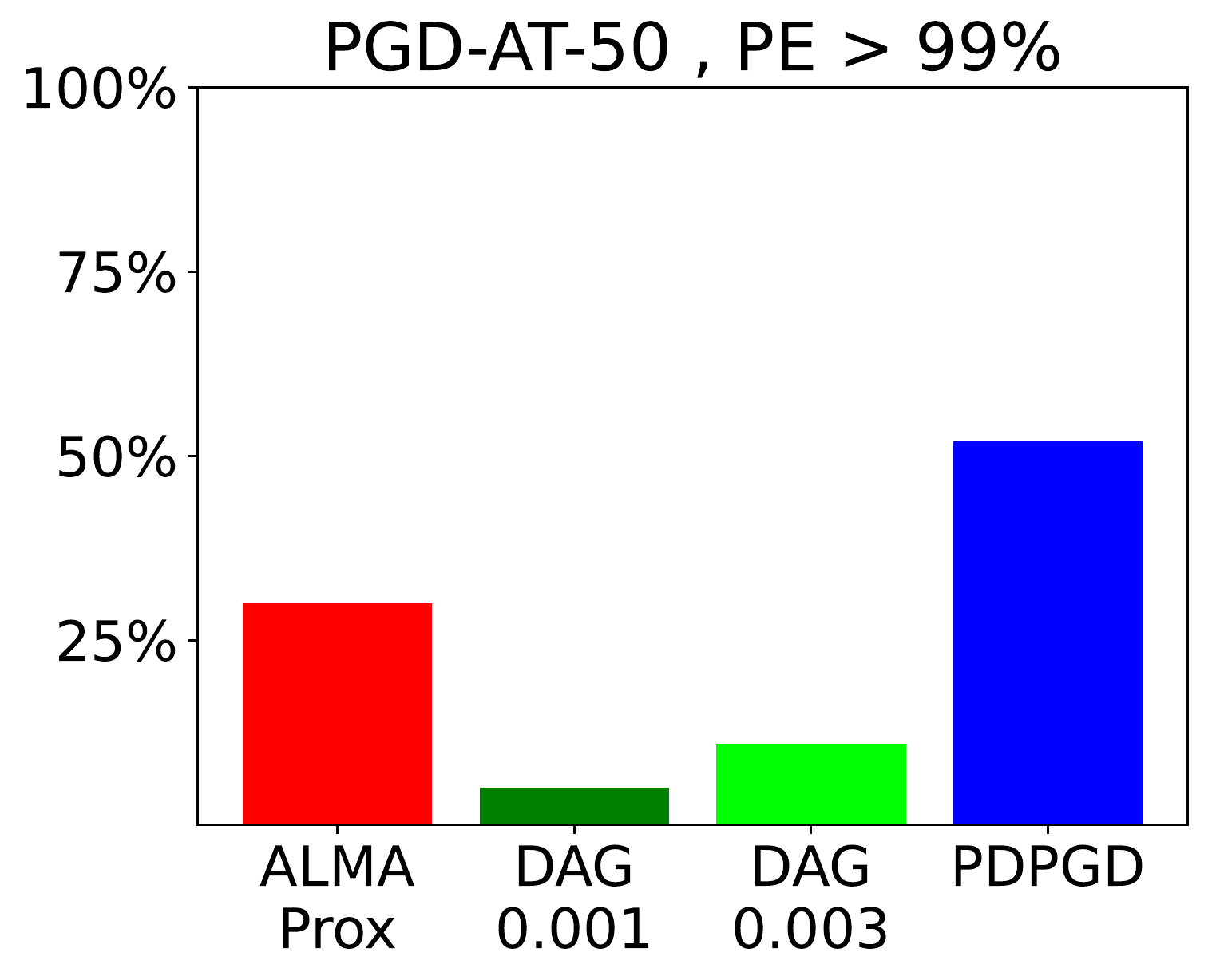}
\includegraphics[width=0.195\columnwidth]{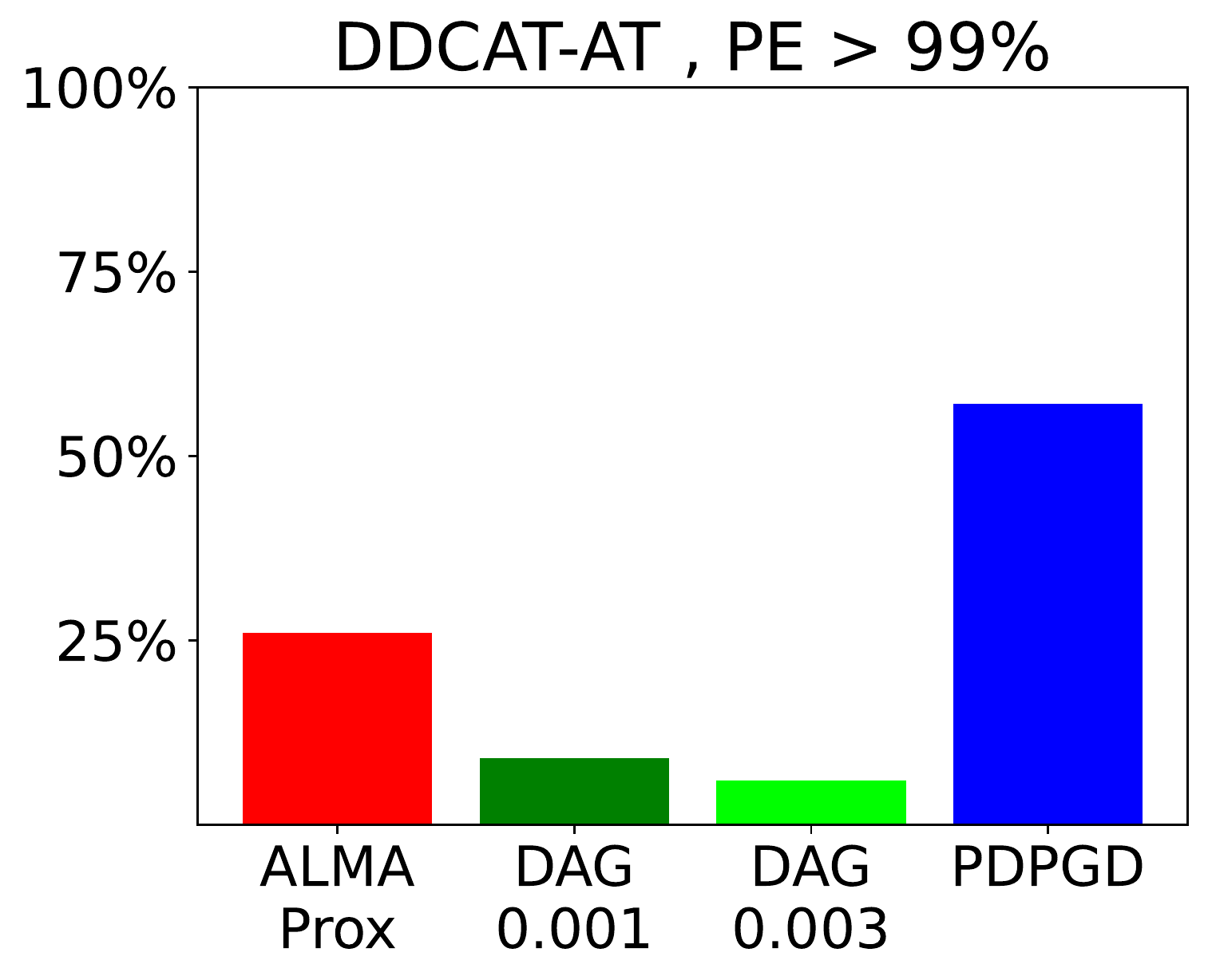}
\includegraphics[width=0.195\columnwidth]{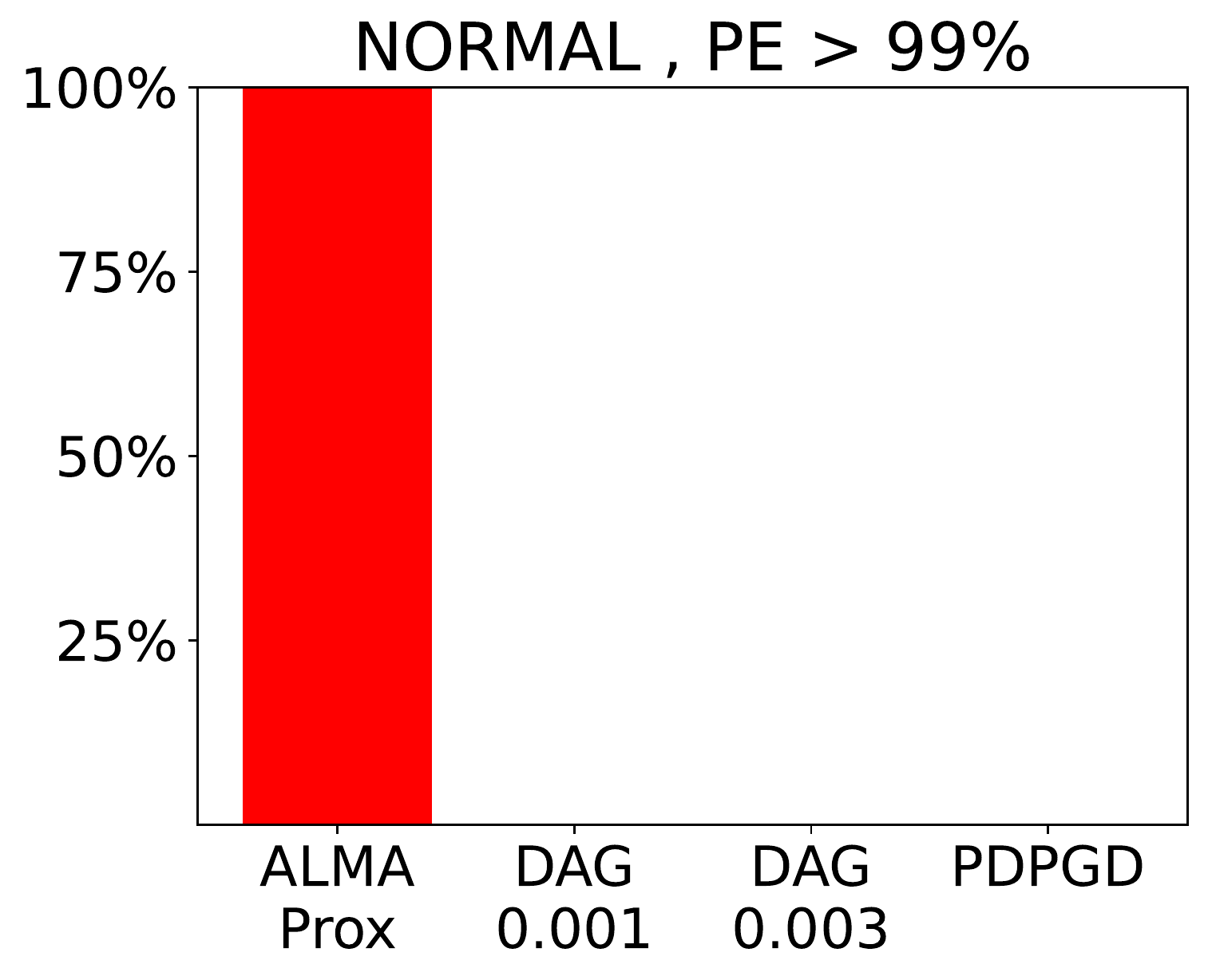}
\caption{Distributions of attack-contributions to the aggregated minimum shown in \cref{fig:pspnet-merge}.}
\label{fig:histogram}
\end{figure}

Although this claim was supported by the results in \cref{fig:pspnet-apsr}, here,
we present additional evidence based on the minimal perturbation
attack experiments in \cref{sec:almaprox}.
In \cref{fig:histogram} we show the distribution of the most successful attacks.
We obtain this distribution via recording for each individual test example the attack
that resulted in the smallest perturbation on that particular example.
We then indicate the proportion of each attack over the test set.

From the distributions it is evident that the different model instances might indeed
be vulnerable to different attacks.
The most successful attacks are ALMAProx and PDPGD overall, but on DDC-AT and PGD-AT-50 the DAG attacks
also contribute significantly to the upper bounds shown in \cref{fig:pspnet-merge}.

\graphicspath{ {Charts/Images} }

\end{document}